\documentstyle[12pt]{article}
\renewcommand{\baselinestretch}{1.5}
\newcommand{\ee }{^{\mbox{\scriptsize EE}}}
\newcommand{\ei }{^{\mbox{\scriptsize EI}}}
\newcommand{\ie }{^{\mbox{\scriptsize IE}}}
\newcommand{\ii }{^{\mbox{\scriptsize II}}}

\newcommand{\E }{^{\mbox{\scriptsize E}}}
\newcommand{\I }{^{\mbox{\scriptsize I}}}

\newcommand{\sn }{_{\mbox{\scriptsize sn}}}
\newcommand{\hopf }{_{\mbox{\scriptsize hopf}}}

\newcommand{\eqdef}{\stackrel{\rm def}{=}}

\begin{document}

\input{psfig}

\begin{titlepage}
\centerline{\huge COVARIANCE PLASTICITY}
\vglue .5cm
\centerline{\huge AND REGULATED CRITICALITY}
\footnotetext
{Supported by the Jean and H\'{e}l\`{e}ne Alfassa fund
for research in Artificial Intelligence,
Office of Naval Research contract N00014-91-J-1021,
National Science Foundation contract DMS-9217655,
and ARL contract MDA972-93-1-0012.}
\vglue 1cm
\centerline{\bf Elie Bienenstock}
\vglue .2cm
\centerline{Division of Applied Mathematics}
\centerline{Brown University}
\centerline{Providence RI 02912}
\centerline{USA}
\centerline{and CNRS, Paris, FRANCE}
\vglue .2cm
\centerline{elie@dam.brown.edu}
\vglue .5cm
\centerline{\bf Daniel Lehmann}
\vglue .2cm
\centerline{Department of Computer Science}
\centerline{Hebrew University}
\centerline{Jerusalem}
\centerline{ISRAEL}
\vglue .2cm
\centerline{lehmann@cs.huji.ac.il}
\vglue .5cm
\centerline{January 1995}

\end{titlepage}

\vglue 1cm
\begin{abstract}
We propose that a regulation mechanism
based on Hebbian covariance plasticity
may cause the brain to operate near criticality.
We analyze the effect of such a regulation
on the dynamics of a network with excitatory and inhibitory neurons
and uniform connectivity within and across the two populations.
We show that, under broad conditions,
the system converges to a critical state
lying at the common boundary of three regions in parameter space;
these correspond to three modes of behavior:
high activity, low activity, oscillation.
\end{abstract}

\newpage

\section{Introduction}
\label{intro}

That evolved brains are highly sensitive organs is an everyday observation.
Viewed as a dynamical system,
a brain may be said to be {\em unusually} susceptible to perturbations
and initial conditions.
This leads one to ask whether
brains may be operating near some form of instability, or criticality,
a hypothesis related to the notions
of computation at the edge of chaos (Langton 1990)
and self-organized criticality (Bak et al. 1987).
In this paper we propose that
while most regulation mechanisms at work in the brain
act according to a classical homeostasis schema,
i.e., have a stabilizing effect,
an opposite effect could result from the regulation of synaptic weights
by a specific form of Hebbian covariance plasticity.
Such a regulation may bring the system near criticality.
We suggest that regulated criticality may be the mechanism
whereby sensitivity is {\em maintained} throughout life
in the face of ongoing changes in brain connectivity.

Hebbian synaptic plasticity (Hebb 1949) plays an important role
in the development of the nervous system,
and is also believed to underlie many instances of learning in the adult.
A {\em covariance rule} of Hebbian plasticity roughly states
that the change in the efficacy of a given synapse
varies in proportion to the covariance
between the presynaptic and postsynaptic activities.
As noted by many authors
(e.g. Sejnowski 1977a, 1977b; Bienenstock et al. 1982;
Linsker 1986; Sejnowski et al. 1988),
a covariance-type rule is preferable to a rule that uses the mere product
of pre- and post-synaptic activities
because the covariance rule predicts not only weight increases
but also activity-related weight decreases,
and as a consequence allows convergence to non-trivial connectivity states.
Some forms of covariance plasticity have been shown to be optimal
for information storage
(Willshaw and Dayan 1990; Dayan and Willshaw 1991; Dayan and Sejnowski 1993).
Also, evidence for Hebbian plasticity of the covariance type
has been reported in several preparations
(Fr\'egnac et al. 1988, 1992; Stanton and Sejnowski 1989;
Artola et al. 1990; Dudek and Bear 1992).

We shall investigate,
in a simple network including excitatory and inhibitory neurons,
the effect of covariance plasticity
acting as a mechanism of {\em regulation,} rather than supervised learning.
Synaptic modification results
in changes---quantitative or qualitative---in the activity
that reverberates in the network;
these changes in turn cause further modification of the weights,
thereby creating a feedback loop between activity and connectivity.
Studying this loop as such,
i.e., independently from any input and output,
we demonstrate that, under fairly general conditions,
it causes the network to converge to a critical surface in parameter space,
the locus of an abrupt transition between different activity modes.
In Metzger and Lehmann (1990, 1994) a similar Hebbian rule
has been studied in the context of supervised learning of temporal sequences.

Schematically, the convergence to a critical state can be explained as follows.
One mode of behavior of a network
including excitatory and inhibitory neurons is oscillation;
such behavior takes place if the synaptic weights
linking excitatory neurons to each other---we
will refer to these as E-to-E weights---are
high enough but not too high.
Oscillation entails high covariance values,
hence, according to the covariance rule,
results in further increase of the E-to-E weights.
If however these weights are higher than a certain critical value---which
depends on other parameters of the system---oscillatory
behavior is impossible,
hence covariance is low or zero,
hence, in accord with the covariance rule used,
the E-to-E weights {\em decrease.}
As a result, the E-to-E weights stabilize around
the critical surface that separates the region of oscillation
from the region(s) of steady firing.

Our study is conducted in the simplest type of network
that will support oscillatory activity:
all synaptic weights of a given type are given identical values,
and so are all firing thresholds of a given type.
This results in a system with just six parameters---four
synaptic weights and two thresholds---and a limited range of behaviors.
Essentially, all neurons fire uniformly, either at a constant rate
(the number of possible rates of firing is one or two, depending on parameters)
or periodically in time.
In the {\em thermodynamic,} i.e., large-size, limit,
the dynamics of the network is adequately described
by a system of differential equations
obtained through a classical mean-field approximation.

We first perform a simple bifurcation analysis of this differential system
(Guckenheimer and Holmes 1983).
We then show that the effect of covariance regulation
is to stabilize the parameter state
at a surface of transition,
where the dynamics exhibits an instability.
Such a critical parameter state for a dynamical system
may be characterized as {\em degenerate.}
A generic, i.e., non-exceptional, state is one
where one expects to find the system
in the absence of further assumptions.
Mathematically, a generic state of a dynamical system is in the {\em interior}
of a parameter region corresponding to a given behavior,
and the system in such a parameter state
is said to be {\em structurally stable;}
the set of non-generic parameter states has measure zero.
We shall show that a state of higher degeneracy,
characterized as a point of intersection of {\em several} critical surfaces,
can be achieved by the simultaneous regulation of {\em several} parameters.
In the vicinity of that highly degenerate state,
the system displays a range of behaviors, including chaos.

The plan of the paper is as follows.
In the next section we study the dynamical properties
of our simple network---in the differential-equation formulation---with
{\em fixed} parameters (synaptic weights and firing thresholds).
We characterize the bifurcations
which take place at the boundaries between domains
corresponding to different modes of behavior.
This study is conducted for a {\em reduced} system,
where the thresholds are eliminated in such a way
as to render the dynamics symmetric about the origin.
Section \ref{regulation} describes the regulation equations.
Section \ref{reduced} describes the behavior of the regulated reduced system,
and Section \ref{full} that of the regulated full system.

\section{The fixed-parameter model}
\label{model}

This section describes the dynamics of the model with fixed parameters.
We first briefly describe a network consisting of a large number ($2N$)
of binary-valued neurons operating under a stochastic dynamics.
However, rather than using this network for our study of plasticity,
we make a number of simplifications and approximations,
leading to a deterministic two-variable differential system
with just six parameters.
The two variables are the excitatory and inhibitory
population averages of cell activity
in the $2N$-dimensional model;
the six parameters include the four average weights of the synapses
within and between these two populations,
as well as the average firing thresholds for the two populations.
We then study the asymptotic behavior
of this differential system for various parameter values.
Different types of asymptotic behavior,
in different regions of the parameter space,
correspond to different {\em phases} of the stochastic system,
and we pay particular attention to the {\em bifurcations} of the solutions,
where the bifurcation parameters are the synaptic weights---see
Schuster and Wagner (1990) and Borisyuk and Kirillov (1992)
for a related bifurcation analysis.
Bifurcations correspond to {\em phase transitions}
in the statistical-physics formulation
(the original $2N$-dimensional model).

We consider a fully-connected network
of $N$ excitatory and $N$ inhibitory linear-sigmoidal
$\{0,1\}$-valued neurons,\footnote
{It is not essential
that the numbers of excitatory and inhibitory neurons be the same.}
operating under a stochastic dynamics.
We denote the activity of the $i$-th excitatory,
resp. inhibitory, neuron by
$x_i\E (t)$, resp. $x_i\I (t)$,
with $x_i\E (t), x_i\I (t)\in\{0,1\}$, $i=1\ldots N$,
and we denote the synaptic weights by
$w_{ij}\ee , w_{ij}\ei , w_{ij}\ie , w_{ij}\ii $, $i,j=1\ldots N$,
where $i$ is postsynaptic and $j$ presynaptic,
and the superscripts indicate,
for each of the two neurons, whether it is excitatory or inhibitory.
Thus, for all $i$ and $j$, $w_{ij}\ee $ and $w_{ij}\ie $ are positive or zero,
whereas $w_{ij}\ei $ and $w_{ij}\ii $ are negative or zero.

The {\em local field} on excitatory neuron $i$,
i.e., the difference between its membrane potential
and its firing threshold $h_i\E $,
is $g_i\E (t)=\sum_j w_{ij}\ee  x_j\E (t) +
\sum_j w_{ij}\ei  x_j\I (t) - h_i\E $.  
Similarly, the local field on inhibitory neuron $j$ is
$g_i\I (t)=\sum_j w_{ij}\ie  x_j\E (t) +
\sum_j w_{ij}\ii  x_j\I (t) - h_i\I $,
where $h_i\I $ is the threshold of inhibitory neuron $i$.
The network dynamics is defined by:
(i) selecting at random, with uniform probability,
one of the $2N$ neurons;
(ii) computing its local field $g(t)$,
of the form $g_i\E (t)$ or $g_i\I (t)$; and
(iii) defining the state of the network at time $t+\delta t$
to be equal to the state at time $t$
except, possibly, for the selected neuron, whose state becomes---or stays---1
with probability $\frac{1}{2}(1+\tanh(\beta g(t)))$.
Parameter $\beta$ is a fixed non-negative number,
an {\em inverse temperature}.
The temperature $T=1/\beta$ measures the amount of noise in the system:
the higher the temperature, the noisier the dynamics.
The update interval is $\delta t = 1/(2N)$,
so that each neuron is updated on average once every time unit.
This {\em asynchronous} dynamics, of the Glauber type (Glauber 1963),
is widely used in statistical-mechanics models;
it lends itself to a convenient mean-field approximation (see below).

A system such as the one just described
will exhibit a highly diverse range of behaviors,
depending on the values of the synaptic weights and firing thresholds.
But we now make the much simplifying assumption that synaptic weights
and firing thresholds are {\em uniform} across each class.
Specifically, for all $i,j=1,\ldots,N$,
we assume that $h_i\E =h\E $, $h_i\I =h\I $,
$w_{ij}\ee =w\ee /N$, $w_{ij}\ei =-w\ei /N$,
$w_{ij}\ie =w\ie /N$, and $w_{ij}\ii =-w\ii /N$,
where $h\E $, $h\I $, $w\ee $, $w\ei $, $w\ie $ and $w\ii $ are fixed parameters,
and $w\ee $, $w\ei $, $w\ie $ and $w\ii $ are non-negative.
The dynamics is thus parameterized by six constants,
four synaptic weights and two thresholds;
$\beta$ is a mere multiplicative factor common to all six parameters,
yet it is convenient to use it as a seventh parameter.
Unless otherwise mentioned, $\beta$ will be 1.

Due to this uniformity assumption,
all neurons in any of the two populations experience the same field
at any given time.
This system exhibits a limited number of fairly simple behaviors,
of which Figure 1 is an example.
This figure shows the time variation of
$\langle x_i\E (t)\rangle$ and $\langle x_i\I (t)\rangle$,
the {\em average} activation levels across
the excitatory and inhibitory populations.
In this example, parameters are:
$N=70$, $w\ee =12$, $w\ie =8$, $w\ei =10$, $w\ii =2$,
$h\E = 1$, $h\I = 3$.
One unit on the time axis corresponds to $2N$ updates,
so that each neuron is updated, on average, once every time unit.
For these parameter values, the system {\em oscillates.}
Note that the oscillation is not perfectly regular, a finite-size effect.
Note also that the inhibitory activity lags somewhat behind the excitatory activity:
the excitatory neurons first trigger the inhibitory ones,
which in turn extinguish, for a while, the excitatory population.
\vglue .4cm
\centerline{\em (Insert Figure 1 around here)}
\vglue .4cm
The presence of oscillations and the amplitude and shape of the waveform
depend on the various parameters.
However, rather than pursuing this study of the stochastic system,
we shall consider the approximation that obtains
in the {\em thermodynamic limit},
that is, when $N\rightarrow\infty$.
The update interval $\delta t=1/(2N)$ then goes to 0
and so does each individual synaptic weight.
Straightforward approximations (Rubin 1988; Schuster and Wagner 1990)
then lead to a continuous-time differential system
for the population averages of the excitatory and inhibitory activation levels,
which we denote, respectively, by $s$ and $\sigma$:
\begin{equation} \left\{ \begin{array}{l}
\dot{s}(t) = .5 - s(t) + .5\tanh[\beta(w\ee s(t)-w\ei \sigma(t)-h\E )]\\
\dot{\sigma}(t)=.5- \sigma(t) +.5\tanh[\beta(w\ie s(t)-w\ii \sigma(t)-h\I )].
\end{array} \right. \label{sys_full} \end{equation}
Note that the variables $s(t)$ and $\sigma(t)$
remain at all $t$ within the interval [0,1].
When $\beta=0$
system \ref{sys_full} has a unique attractor, $(s,\sigma)=(.5,.5)$.
Indeed, in the high-temperature limit,
all neurons act independently of each other
and fire with probability .5 at each time.

We shall now make a last simplification,
whose purpose it is to render $(.5,.5)$ a fixed point---though
not necessarily stable---at {\em all} temperatures
and for all values of the synaptic weights.
This is easily achieved by letting the thresholds $h\E $ and $h\I $
be determined by the synaptic weights as follows:
\begin{equation} \begin{array}{l}
h\E  = .5 (w\ee -w\ei )\\
h\I  = .5 (w\ie -w\ii ).
\end{array} \label{chvar} \end{equation}
It is then convenient to adopt the change of variables:
$s\mapsto s-.5$, $\sigma\mapsto \sigma-.5$,
and system \ref{sys_full} becomes:
\begin{equation} \left\{ \begin{array}{l}
\dot{s}(t) = -s(t) + .5\tanh[\beta(w\ee s(t)-w\ei \sigma(t))]\\
\dot{\sigma}(t) = -\sigma(t) + .5\tanh[\beta(w\ie s(t)-w\ii \sigma(t))].
\end{array} \right. \label{sys_red} \end{equation}
In \ref{sys_red}, the variables $s$ and $\sigma$
are in the interval $[-.5,+.5]$,
and the only parameters left are the four synaptic weights
and the inverse temperature.
For all parameter values, the origin is a fixed point of system \ref{sys_red}.
A different position for the fixed point
could be obtained with an appropriate modification of equations \ref{chvar},
yet in the current version the fixed point is also a center of symmetry.
For the moment,
this hard-wired symmetry should be regarded as an {\em ad-hoc} device,
whose purpose is to make the analysis more convenient.
We shall refer to system \ref{sys_full} as the {\em full} system,
and to system \ref{sys_red} as the {\em reduced} system.
We shall see in Section \ref{full}
that under appropriate regulation
the two systems behave similarly.

We now discuss some important properties of the reduced system,
system \ref{sys_red} (see also Rubin 1988).
Consider first Figure 2a (phase diagram), which shows
four trajectories of the state $(s(t),\sigma(t))$;
the starting points of these trajectories are indicated by triangles.
The parameters (synaptic weights) used in this example are identical
to those used in Figure 1, i.e.,
$w\ee =12$, $w\ie =8$, $w\ei =10$, $w\ii =2$.
As expected, the asymptotic behavior is {\em periodic;}
there is a limit cycle which attracts all points of the square $[-.5,.5]^2$,
except the unstable equilibrium $(0,0)$.
Motion is counterclockwise,
for, as mentioned above, $\sigma(t)$ lags behind $s(t)$.

In addition to these four orbits, Figure 2a shows two curves,
the $s$- and $\sigma$-{\em nullclines} for system \ref{sys_red}.
These are the loci of the points $(s,\sigma)$ such that $ds/dt$,
resp. $d\sigma/dt$, vanish.
The equations for the $s$- and $\sigma$-nullclines
are easily seen to be, respectively:
\begin{equation}
\sigma = \frac{1}{w\ei }(w\ee s - T\tanh^{-1}(2s)),
\label{snull}\end{equation}
\begin{equation}
s = \frac{1}{w\ie }(w\ii \sigma + T\tanh^{-1}(2\sigma)).
\label{signull}\end{equation}
The $\sigma$-nullcline is an increasing sigmoid-shaped curve,
whereas the $s$-nullcline generally has the shape of an `S' lying on its side.
Of particular interest are the intersection points of the two nullclines;
these are the {\em fixed points} of the dynamics.
In the case illustrated in Figure 2a, the only intersection is $(0,0)$,
an unstable equilibrium.
Trajectories intersect the $s$-, resp. $\sigma$-, nullcline
in a direction parallel to the $\sigma$-, resp. $s$-, axis.

The study of the nullclines is of interest
because it is often possible
to predict how a parameter change will affect the dynamics of the system
by reasoning about how the nullcline diagram will change;
the bifurcation we shall be mostly interested in
is associated with a conspicuous change in this diagram.
Note that the $s$-nullcline is affected by parameters $w\ee $ and $w\ei $,
whereas the $\sigma$-nullcline is affected by parameters $w\ii $ and $w\ie $.
\vglue .4cm
\centerline{\em (Insert Figure 2 around here)}
\vglue .4cm
Let us consider first the changes brought about
by letting parameter $w\ee $ grow,
starting from the point $w\ee  = 12$
for which the system oscillates;
other parameters are unchanged.
When $w\ee $ grows, the slope of the central,
quasi-linear, part of the $s$-nullcline increases
(see equation \ref{snull});
that part of the curve rotates about the symmetry center (0,0).
As a result, the peak of the $s$-nullcline to the right approaches
the upper part of the sigmoid-shaped $\sigma$-nullcline,
and the minimum of the $s$-nullcline to the left approaches
the lower part of the $\sigma$-nullcline.
Eventually, at a certain critical value $\hat w\ee \sn $
(subscript `sn' stands for `saddlenode'---see below),
the two curves become tangent to each other.
This happens in two points at once,
near the upper right-hand corner and near the lower left-hand corner,
due to the symmetry of the system.
This situation is depicted in Figure 2b:
$w\ee $ is exactly equal to the critical value $\hat w\ee \sn $
(with parameters as above, $\hat w\ee \sn \approx 14.22$),
and the nullclines are just tangent to each other.

When $w\ee $ grows a little further,
each point of contact splits into two intersection points,
of which one is an attractor.
Figure 2c shows this situation, with $w\ee =15$,
somewhat above the critical value $\hat w\ee \sn $.
Four trajectories are shown, in addition to the two nullclines.
The system has five fixed points,
three unstable ones and two stable ones (attractors).
Only the stable fixed points are of interest to us;
they are very near the upper right-hand
and lower left-hand corners of the square,
corresponding to high,
respectively low, excitatory and inhibitory activities.

The bifurcation occurring at $\hat w\ee \sn $
is of the {\em saddlenode} type.
It results in a drastic change of behavior of the system:
the periodic attractor disappears
and is `siphoned' into the two new point attractors.
These two points attract the entire square,
except a set of measure 0
which includes the three unstable fixed points.
Thus, altough this bifurcation
is caused by a mere {\em local} change,
namely the intersection of the nullclines,
it results in a reorganization of the dynamics
that is both abrupt and {\em global.}\footnote
{As mentioned, {\em two} distinct saddlenode bifurcations
take place simultaneously.
Such a double bifurcation is not generic;
it occurs here due to the symmetry that we introduced
when reducing system \ref{sys_full} into system \ref{sys_red}.}

Having described the breakdown of oscillations
when parameter $w\ee $ is increased,
we now consider the opposite change, that is,
we let $w\ee $ decrease.
This results in a decrease of the slope of the central,
increasing, portion of the $s$-nullcline (equation \ref{snull}).
Eventually, the curve becomes monotonically decreasing;
this does not alter the number of intersections of the nullclines,
point $(0,0)$ remaining the sole equilibrium.
However, the amplitude of the limit cycle decreases along with $w\ee $.
The cycle eventually collapses to a point;
the equilibrium $(0,0)$ has then become stable.
This can be seen in a linear stability analysis
of system \ref{sys_red} around point $(0,0)$.
It is easily shown that,
in case there are two complex conjugate eigenvalues,\footnote
{The condition for this is $4w\ei w\ie >(w\ee +w\ii )^2$.}
the real part of these eigenvalues is negative if and only if $w\ee <w\ii +4T$.
Thus, $w\ii +4T$ is a critical value for parameter $w\ee $.
We define $\hat w\ee \hopf \eqdef w\ii +4T$
(with the current parameter setting, $\hat w\ee \hopf =6$).
The change of behavior occurring at $\hat w\ee \hopf $
is a {\em normal}\footnote
{That is, supercritical.
However, for very large values of $w\ie $,
the bifurcation is subcritical---see footnote 5.}
Hopf bifurcation.

So far, we studied the behavior of system \ref{sys_red}
for different values of parameter $w\ee $, all other parameters being fixed.
In other words, we described the system's behavior
on a particular 1-dimensional subspace
of the 4-dimensional parameter space.
We now extend this study to a 2-dimensional subspace, the $(w\ee ,w\ie )$ plane.
Figure 3a is the bifurcation diagram of system \ref{sys_red} in that plane,
with other parameters as before ($w\ei =10$, $w\ii =2$).
This diagram shows three distinct regions,
corresponding to three different attractor configurations;
unstable fixed points and unstable limit cycles are ignored in this diagram.
In the middle region---which we call region $\cal P$,
for {\em Periodic}---the system oscillates.
The boundary of this region to the right
is the saddlenode bifurcation curve, which we denote $\cal S$;
as discussed above,
the rightmost region has {\em two} point attractors,
and we call it region $\cal T$.
The leftmost region, which we call $\cal O$,
has only one point attractor, the center of symmetry $(0,0)$;
it is separated from region $\cal P$ by the Hopf bifurcation curve,
a vertical line of equation $w\ee =\hat w\ee \hopf $.
The curve in the lower left of the diagram,
separating region $\cal O$ from region $\cal T$,
is the locus of a {\em pitchfork} bifurcation.
This bifurcation diagram, obtained for one particular set of values
of the parameters $w\ei , w\ii $ and $\beta$,
is representative of the general case.\footnote
{It is however simplified in two ways.
First, the transition from region $\cal P$ to region $\cal T$
is of the saddlenode type only for large enough values of $w\ie $;
this range of values corresponds roughly
to the straight portion of curve $\cal S$ (Figure 3a).
To see why this is so, consider again Figure 2b,
the nullcline diagram at the bifurcation, with $w\ie = 8$.
Note that the points of contact between the nullclines
appear near the corners of the square, far from the origin;
this is due to the fact that $w\ie $ is relatively large,
hence the slope of the $\sigma$-nullcline at the origin
is larger than the slope of the $s$-nullcline.
The bifurcation is then of the saddlenode type, as described.
If however $w\ie $ is small,
hence so is the slope of the $\sigma$-nullcline at the origin, 
the transition from $\cal P$ to $\cal T$ as $w\ee $ is increased
takes place differently.
A pair of intersection points between the nullclines
first split off {\em from the origin;}
these are unstable equilibria.
As $w\ee $ increases,
these two equilibria move away from the origin,
while remaining inside the large stable limit cycle.
At a certain critical value for $w\ee $ they become stable---a
(double) subcritical Hopf bifurcation---and
almost immediately thereafter the large limit cycle disappears.
Thus, the transition from region $\cal P$ to region $\cal T$
really takes place in two steps,
giving rise to a {\em three-attractor} behavior:
the system has one large limit-cycle attractor
{\em as well as} two point attractors,
the latter being inside the cycle.
The region of the $(w\ee ,w\ie )$ plane where this behavior takes place
is a strip extending along the lower, curved,
part of the $\cal P/\cal T$ boundary;
it is too narrow to be seen in Figure 3a.
(With parameters $w\ei $ and $w\ii $ as above and $w\ie =2.75$,
the three-attractor behavior occurs for $w\ee $ between $8.993$ and $9.030$.
For some other values of $w\ei $ and $w\ii $ this behavior does not occur at all,
and the transition from $\cal P$ to $\cal T$ is always of the saddlenode type.)
For the purpose of this paper (see footnotes 8 and 11)
it is important to note that the point attractors
appear either {\em exactly} or {\em almost} at the same time
as the periodic attractor disappears.
The second approximation in the bifurcation diagram,
mentioned only for the sake of completeness,
concerns the $\cal O$-to-$\cal P$ transition.
This is generally a smooth, supercritical, Hopf bifurcation.
However, as mentioned in footnote 4, this Hopf bifurcation becomes subcritical
for very large values of $w\ie $.
There is thus a narrow region to the left of the bifurcation line $w\ee =\hat w\ee \hopf $
where the limit-cycle attractor coexists with the point attractor (0,0);
for instance, at $w\ie =100$, the width of this region is $\approx 0.63$.}
\vglue .4cm
\centerline{\em (Insert Figure 3 around here)}
\vglue .4cm
In sum, the $(w\ee ,w\ie )$ bifurcation diagram for system \ref{sys_red}
is characterized by a central periodic-attractor region,
a large vertical patch extending to $+\infty$ in the $w\ie $ direction
(phase $\cal P$),
flanked by point-attractor regions on each side (phases $\cal O$ and $\cal T$).
The transition from $\cal P$ to $\cal T$ is abrupt ($\cal S$ line),
while the transition from $\cal O$ to $\cal P$ is smooth.
As mentioned in the Introduction, system \ref{sys_full}---the full system---is
not amenable to such a thorough analysis;
however, we shall see in Section \ref{full}
that the two systems behave in much the same way
under the plasticity rules that we shall now introduce.

\section{The regulation equations}
\label{regulation}

Whereas in the previous section the synaptic weights $w\ee $ and $w\ie $
were fixed parameters, they will now be made to evolve.
Their evolution will obey a Hebbian covariance rule, hence be a function
of second-order temporal averages of the dynamic variables $s$ and $\sigma$.
Synaptic plasticity creates a {\em regulation loop:}
changing the parameters affects the dynamics of the system,
which in turn alters the second-order moments of $s$ and $\sigma$.
Formally, the regulation is implemented by
introducing additional differential equations, coupled to system \ref{sys_red}
(or to system \ref{sys_full}---see Section \ref{full}).
The rate of change of $w\ee $ and $w\ie $ will typically be
several orders of magnitude slower than that of $s$ and $\sigma$.

Let us first define, for any function of time $r(t)$,
a moving time average:
\begin{equation}
\bar{r}(t) = \rho \int_{-\infty}^t r(u) e^{\rho (u-t)} du.
\nonumber\end{equation}
Parameter $\rho$ is a positive constant, physically an inverse time;
the larger $\rho$, the narrower the averaging kernel.
Equivalently, $\bar{r}(t)$ may be defined by a differential equation,
more convenient for simulation purposes:
\begin{equation}
\frac{d\bar{r}(t)}{dt} = \rho (r(t) - \bar{r}(t)).
\nonumber\end{equation}
Consider now, with reference to the original stochastic model (Section \ref{model}),
the {\em instantaneous covariance} between two excitatory neurons $i$ and $j$,
defined as:
$c_{ij}\ee (t) \eqdef (x_i\E (t)- \bar x_i\E (t))(x_j\E (t) - \bar x_j\E (t))$.
If we take the {\em population average} $\langle c_{ij}\ee (t)\rangle$
of this instantaneous covariance,
we obtain, in the thermodynamic limit $N\rightarrow\infty$,
the instantaneous variance of $s(t)$:
\begin{equation}
c\ee (t) \eqdef (s(t) -\bar{s}(t))^2.
\label{covee}\end{equation}

It is this quantity $c\ee $ that we use to regulate
the excitatory-to-excitatory synaptic weight $w\ee $.
The regulation equation is linear in $c\ee $:
\begin{equation}
\frac{dw\ee (t)}{dt} = \varepsilon\ee  ( c\ee (t) - \theta\ee  ).
\label{regee}\end{equation}
Parameters $\varepsilon\ee $ and $\theta\ee $ are positive.
Note that the quantity $c\ee (t)$ is always non-negative;
the term $-\theta\ee$ is therefore necessary
to allow for decreases of $w\ee $.

We shall also consider a regulation for $w\ie $,
the synaptic weight from excitatory to inhibitory neurons,
although this regulation will play a less important role than that of $w\ee $. 
The modification rule for $w\ie $ has the same form as equation \ref{regee},
yet it uses the excitatory-to-inhibitory instantaneous covariance,
defined as:
\begin{equation}
c\ie (t) \eqdef (s(t) -\bar{s}(t)) (\sigma(t) -\bar{\sigma}(t)).
\label{covie}\end{equation}
The regulation equation for $w\ie $ then reads:
\begin{equation}
\frac{dw\ie (t)}{dt} = \varepsilon\ie  ( c\ie (t) - \theta\ie  ).
\label{regie}\end{equation}
In equation \ref{regie}, $\theta\ie $ is a positive constant,
as $\theta\ee $ in equation \ref{regee}.
However, the modification rate constant $\varepsilon\ie $ is negative.
The main reason for this will be given in the next section;
for now, note that this choice is consistent with the spirit of Hebb's principle,
for, when considered {\em postsynaptically} to the target neuron,
the effect of synapse reinforcement if that target neuron is inhibitory
is the opposite of the effect obtained if the target neuron is excitatory.

\section{Behavior of the regulated reduced system}
\label{reduced}

This section describes the behavior of the regulated reduced system.
We demonstrate that each of the two regulation loops
introduced in Section \ref{regulation},
when acting separately,
brings the system to the critical surface $\cal S$,
the locus of an abrupt phase transition (saddlenode bifurcation).
We then examine the behavior of the system
with the two regulation loops active simultaneously;
we show that under some conditions
the state converges to a point on $\cal S$
with a remarkable nullcline configuration.

Before we consider the regulation proper,
let us examine how the covariances
change across the $(w\ee ,w\ie )$ plane.
Figure 3b shows the values of $\bar c\ee $,
the time average of the instantaneous variance of $s(t)$,\footnote
{This corresponds, in the original system,
to the population- {\em and} time-average of the covariance,
$\langle \bar c_{ij}\ee \rangle$;
the latter becomes $\bar c\ee $
in the thermodynamic limit $N\rightarrow\infty$.
In the regulation equation,
we use the {\em instantaneous} covariance $c\ee (t)$
rather than its time average $\bar c\ee $ (see Discussion).
The time-averaged variance $\bar c\ee $
is used here for illustration purposes only.
In order to obtain an essentially constant value for $\bar c\ee $
rather than an oscillating function of time,
different values of $\rho$ are used for the two averaging operations:
the kernel used to average $c\ee $ into $\bar c\ee $
is ten times broader than the kernel used to compute $\bar s$ from $s$.}
along several horizontal lines in the $(w\ee ,w\ie )$ plane.
As expected, $\bar c\ee $ is positive only in region $\cal P$,
where the dynamics is periodic;\footnote
{In general, positive average covariance across a neuronal population
indicates collective fluctuations;
in our simplified two-dimensional system,
the only possible nontrivial asymptotic behavior
is periodic oscillation.}
although not shown, the same is true of $\bar c\ie $,
the time average of the E-to-I covariance.
Note that as $w\ee $ crosses the $\cal O$-to-$\cal P$ boundary
(Hopf bifurcation) from left to right,
$\bar c\ee $ increases {\em smoothly} from 0 to positive values:
as discussed above, the amplitude of the limit cycle
at this bifurcation is infinitesimal.
In contrast, the change in $\bar c\ee $ and in $\bar c\ie $
at $\cal S$ (saddlenode bifurcation) is a sharp one,
as the system undergoes there a transition from a {\em large} limit-cycle regime
to a fixed-point attractor.

We now start our study of covariance plasticity
by regulating parameter $w\ee $ in system \ref{sys_red}
while all other parameters, including $w\ie $, remain fixed.
The system under study then consists
of coupled equations \ref{sys_red}, \ref{covee}, \ref{regee}.
Equation \ref{regee} prescribes an increase of $w\ee $ when $c\ee > \theta\ee $,
and a decrease when $c\ee < \theta\ee $.
Referring to Figure 3b, we see that to the left of $\cal S$,
where $c\ee $ is high, the first of the two conditions applies;
in this region $w\ee $ increases.
To the right of $\cal S$ the covariance vanishes, and $w\ee $ decreases.
Therefore, $w\ee (t)$ is attracted to the transition line $\cal S$.\footnote
{The control parameter $\theta\ee $ should be smaller
than the value of $\bar c\ee $ immediately to the left of $\cal S$.
The portion of the boundary line
where the bifurcation is a subcritical Hopf rather than a saddlenode
(footnote 5) yields similar behavior,
since the disruption of the large-amplitude limit cycle
occurs very near the emergence of point attractors (see also footnote 11).}
\vglue .4cm
\centerline{\em (Insert Figure 4 around here)}
\vglue .4cm
The behavior of this $w\ee $ regulation loop is illustrated in Figure 4a
for the following setting of parameters:
$w\ei = 10$, $w\ii = 6$, $\rho=.1$, $\theta\ee =.01$, $\varepsilon\ee =.01$.
This figure focuses on a small region of the $(w\ee ,w\ie )$ plane,
and shows the projection of the trajectory of $(s, \sigma, w\ee ,w\ie )$.
Several trajectories are shown, all horizontal since $w\ie $ is a constant,
These trajectories terminate on the critical line $\cal S$,
and the behavior of the $s$ and $\sigma$ components on them is as follows.
On the trajectories coming from the left, in the $\cal P$ region,
$(s,\sigma)$ moves along a cyclic orbit,
whose amplitude grows as $w\ee $ increases
and approaches the bifurcation line.
On the trajectories coming from the right, in the $\cal T$ region,
$(s, \sigma)$ stays in one of the two point attractors while $w\ee $ decreases
until it reaches the bifurcation curve.
When $\cal S$ is reached, either from the left or from the right,
motion does not really stop.
Rather, $w\ee $ sets in a periodic oscillation of small amplitude
synchronized with a large-amplitude periodic motion of $(s,\sigma)$;
the frequency of this oscillation 
is several orders of magnitude slower than in $\cal P$,
hence covariance is small---it
matches, on average, the control parameter $\theta\ee $.
When in this regime, the system spends a long time in one of the two 
almost-attracting corners of the $[-.5,+.5]^2$ box before leaving it
and moving rapidly to the other corner.
This results in an almost-square wave,
a behavior that is intermediate between the fast periodic motion observed
in $\cal P$ and the bistable situation prevailing in $\cal T$.
The period of this oscillation and the amplitude
of the oscillation of $w\ee $
depend on parameters $\rho$, $\theta\ee $, and $\varepsilon\ee $.\footnote
{Not shown on Figure 4a is the leftmost part of region $\cal P$,
near the Hopf bifurcation,
where the limit cycle is of small amplitude
hence the condition $\bar c\ee  > \theta\ee $ is not realized.
When initialized there,
the system does not converge to $\cal S$.
However, in both the $w\ee $ and the $w\ie $ directions,
the domain of attraction of $\cal S$ extends to $+\infty$.}

We next consider the $w\ie $-regulated system,
where $w\ee $ and all other parameters remain fixed.
This system consists of coupled equations \ref{sys_red}, \ref{covie}, \ref{regie}.
As noted, the E-to-I covariance $c\ie $ vanishes
outside region $\cal P$, just like $c\ee $;
within $\cal P$ it varies, in a first approximation, like $c\ee $.
Since we chose $\varepsilon\ie $ to be {\em negative,}
$w\ie $ decreases in $\cal P$ and increases in $\cal T$,
whereas the opposite was true of $w\ee $
when it was regulated.
Figure 4b shows this $w\ie $ dynamics
in the same region of the $(w\ee ,w\ie )$ plane as before.
Parameters are $w\ei = 10$, $w\ii = 6$, $\rho=.1$,
$\theta\ie =.01$ and $\varepsilon\ie =-.01$.
The trajectories are now parallel to the $w\ie $ axis,
and $(w\ee ,w\ie )$ is again attracted to the critical line $\cal S$
separating region $\cal P$ from region $\cal T$.
This is true only to the left of the vertical asymptote of that curve;
trajectories to the right of that line go to $+\infty$.

In sum, regulation of either one of the two parameters $w\ee $, $w\ie $
has the effect of bringing the system
to the critical surface $\cal S$ separating the region of oscillation
from the region of bistable steady firing;
the nullcline diagram is then as in Figure 2b.
Note that when the system is on $\cal S$,
a small perturbation in the weights will elicit either oscillation,
constant firing at near-maximum rate,
or constant firing at near-minimum rate.

We now turn to the behavior of the system
when the two regulation loops act simultaneously;
we thus study the system of coupled equations
\ref{sys_red}, \ref{covee}, \ref{regee}, \ref{covie}, \ref{regie}.
Figure 4c shows the $(w\ee ,w\ie )$ dynamics for the same parameters as before,
i.e., $w\ei = 10$, $w\ii = 6$, $\rho=.1$, $\theta\ee =.01$, $\varepsilon\ee =.01$,
$\theta\ie =.01$ and $\varepsilon\ie =-.01$.
It appears from this diagram that the evolution proceeds
in two clearly distinct stages.
In the first stage,
which could be predicted from the study of the regulation loops
acting separately,
$(w\ee ,w\ie )$ moves toward line $\cal S$.\footnote
{The direction of this linear motion is roughly parallel to the line $w\ee = -w\ie $.
This is because $\varepsilon\ee =-\varepsilon\ie $,
$\theta\ee = \theta\ie $,
and the two covariances $c\ee $ and $c\ie $ are nearly the same.
Another choice of parameters would result in a different slope,
but otherwise similar behavior.}
When this line is reached,
motion slows down considerably---typically
by several orders of magnitude---and
proceeds {\em along} the critical line,
eventually converging to a point on $\cal S$ denoted $G$ in Figure 4c.
As before, attractor $G$ is in reality a slow limit cycle,
of small amplitude in $w\ee $ and $w\ie $,
and large amplitude in $s$ and $\sigma$.
All four variables, $s(t)$, $\sigma(t)$, $w\ee (t)$, and $w\ie (t)$,
are now synchronized;
the distinction between slow and fast variables has thus vanished.
The basin of attraction of $G$ in the $(w\ee ,w\ie )$ plane
roughly consists of the {\em union} of the two domains of attraction
of $\cal S$ for the separate $w\ee $ and $w\ie $ regulation dynamics;
only the region to the left of and around the Hopf line
is not attracted to the saddlenode line $\cal S$ and eventually to $G$.
\vglue .4cm
\centerline{\em (Insert Figure 5 around here)}
\vglue .4cm
The location of $\cal S$ in the $(w\ee ,w\ie )$ plane depends
on the values of the fixed parameters $w\ei $ and $w\ii $.
The location of the attractor $G$ on $\cal S$
further depends on the control parameters $\theta\ee $ and $\theta\ie $.
When the latter are given identical values,
as in the case illustrated in Figure 4c,
the attractor $G$ has the remarkable property
that the $s$- and $\sigma$-nullclines stand in {\em near overlap}
over a large portion of the interval [-.5,+.5] (Figure 5);
the flow of the system in this configuration
nearly vanishes on a large one-dimensional manifold
in the two-dimensional phase space.
Further, $s(t)$ and $\sigma(t)$ remain nearly identical at all times.\footnote
{Giving different values to parameters $\theta\ee $ and $\theta\ie $
mostly affects the behavior of the system
after it has reached $\cal S$;
if $\theta\ee $ is larger, resp. smaller, than $\theta\ie $,
the state moves downward, resp. upward, on $\cal S$.
When $(w\ee ,w\ie )$ is on $\cal S$ but above point $G$,
the nullclines are tangent to each other but do not overlap;
such a situation is illustrated in Figure 2b.
When $(w\ee ,w\ie )$ is on $\cal S$ but below point $G$,
the nullclines do overlap, but over a smaller domain.
With $\theta\ee = .0118$ and $\theta\ie = .0100$,
the state stabilizes in the narrow three-attractor region described in footnote 5.
The state $(s,\sigma)$ then visits each of the three `attractors' in turn:
its motion consists of a succession
of large-amplitude oscillations (periodic attractor)
and of spiraling orbits around two symmetric points
in the interior of the large cycle (point attractors).
The amplitude of the motion of $(w\ee ,w\ie )$ remains small.
This is a mildly chaotic behavior;
a more pronounced chaotic behavior
will be described in the next section for the full system.}

\section{Behavior of the regulated full system}
\label{full}

Recall that system \ref{sys_red},
which we used so far,
was derived from system \ref{sys_full}
by eliminating the firing thresholds $h\E $ and $h\I $ (equations \ref{chvar})
in such a way as to make $(.5,.5)$---$(0,0)$ in system \ref{sys_red}---a
center of symmetry of the dynamics.
While easier to analyze,
the reduced system is less realistic.
There is no clear biological justification for this hard-wired symmetry;
moreover, when the system is in phase $\cal T$,
i.e., to the right of the critical surface $\cal S$,
it can stay for arbitrarily long periods of time in one
of the two fixed point attractors, e.g. in the high-activity one;
this is unrealistic.

In this section we consider a biologically more plausible way
of introducing symmetry in the dynamics.
Rather than eliminating the thresholds according to equations \ref{chvar},
we {\em regulate} them,
thereby implementing a form of `soft' symmetry.
Regulating the firing thresholds in a neural network
is a simple way to maintain the mean activity
around an intermediate, useful, value.
This may be viewed as a simplification of the regulation mechanisms
at work in real brains,
which, in all likelihood, involve systems of inhibitory neurons
acting on various time scales.

The simultaneous regulation of four parameters
results in a complex dynamics,
which makes a thorough analysis impractical.
We shall proceed as follows.
We first consider, in system \ref{sys_full},
the regulation of $w\ee $ and $h\E $
for a given setting of all other parameters.
We show that the system converges to the intersection of two critical curves,
each of which corresponds to the establishement
of one point of contact between the nullclines.
We next consider the system
with all four parameters $h\E $, $h\I $, $w\ee $ and $w\ie $ regulated,
and study the projection of the dynamics on the $(w\ee ,w\ie )$ plane.
There are again two stages;
the first essentially reproduces the behavior observed
with the sole $(w\ee ,h\E )$ regulation,
while the second is analogous to that observed
when regulating $w\ee $ and $w\ie $ in the reduced system;
this applies for a broad range of the remaining fixed parameters $w\ei $ and $w\ii $.

Figure 6a is the bifurcation diagram of system \ref{sys_full}
in the $(w\ee ,h\E )$ plane,
for the following values of the fixed parameters:
$w\ei =10$, $w\ie =10$, $w\ii =1$, $h\I =5$.
As before, we ignore unstable equilibria and unstable limit cycles.
As before there are three regions,
denoted respectively by $\cal O$, $\cal T$ and $\cal P$,
corresponding to three types of asymptotic behavior:
single fixed-point attractor;
two fixed-point attractors (high and low activity);
one periodic attractor.
We now however subdivide region $\cal O$---somewhat arbitrarily---according
to the location of the fixed-point attractor in the phase space:
the three subregions denoted ${\cal O}_h$, ${\cal O}_m$, and ${\cal O}_l$,
correspond, respectively, to high, middle, and low activity for this attractor.
The transition between region $\cal P$ and region ${\cal O}_m$
takes place through the familiar, smooth, Hopf bifurcation.
The transition between ${\cal P}$ and ${\cal O}_h$,
as well as its continuation between ${\cal O}_l$ and $\cal T$,
takes place through a saddlenode bifurcation.
We denote by ${\cal S}_h$ the locus of this transition;
it marks the appearance of a point of contact between the nullclines
near the high-activity corner,
and is thus similar to the $\cal S$ transition in the reduced system.
However, due to the symmetry of that system,
another point of contact appeared simultaneously near the low-activity corner,
giving rise to a double bifurcation.
In system \ref{sys_full} this is no more the case,
and the intersection of the nullclines near the low-activity corner
gives rise to a distinct saddlenode bifurcation line,
the transition between ${\cal P}$ and ${\cal O}_l$,
which we denote ${\cal S}_l$.
\vglue .4cm
\centerline{\em (Insert Figure 6 around here)}
\vglue .4cm
When regulating $w\ee $ according to equation \ref{regee}
and leaving all other parameters fixed,
the behavior of system \ref{sys_full} is as follows.
When starting in region $\cal P$ to the left of the critical line ${\cal S}_h$,
the system oscillates, covariance is high, hence $w\ee $ increases
until it reaches the critical line ${\cal S}_h$.
A point of contact is then established
near the high-activity corner of the square.
The system settles in a slow periodic attractor,
of small amlitude in $w\ee $ and large amplitude in $(s,\sigma)$,
whereby nearly all the time is spent in the high-activity state.

We now regulate the threshold $h\E $ as well,
in such a way as to stabilize $\bar{s}$,
the time average of $s$, around a given target value $\theta\E $:
\begin{equation}
\frac{dh\E (t)}{dt} = \varepsilon\E ( \bar{s}(t) - \theta\E  ).
\label{reghe}
\end{equation}
The rate constant $\varepsilon\E$ is positive and small,
and the control parameter $\theta\E $ is chosen
well in the interior of the interval $(0,1)$,
e.g. between .2 and .8
(remember that in system \ref{sys_full}
the activity variables $s$ and $\sigma$ lie in the interval $(0,1)$).
To see how equation \ref{reghe} achieves the desired regulation,
note for instance that, if $\bar{s}(t) > \theta\E $, $h\E $ will increase,
which in turn will result in a decrease of $\bar{s}(t)$.

When both $w\ee $ and $h\E $ are regulated,
the system converges to the {\em intersection}
of the two critical lines ${\cal S}_h$ and ${\cal S}_l$.
In effect, we saw that the full system,
when at a generic point of ${\cal S}_h$,
stays nearly all the time in the high-activity state;
this results in a high value of $\bar {s}$.
To achieve the condition $\bar {s} \approx \theta\E $,
the equilibrium for equation \ref{reghe},
the system can only be on ${\cal S}_l$ {\em as well.}

The joint $(w\ee ,h\E )$ dynamics is illustrated in Figure 6b,
for parameters $w\ei $, $w\ie $, $w\ii $ and $h\I $ as above,
and $\rho = .2$, $\varepsilon\E = .001$, $\theta\E = .5$,
$\varepsilon\ee =.01$, $\theta\ee =.01$.
The intersection of ${\cal S}_h$ and ${\cal S}_l$, denoted $F$ in Figure 6b,
is reached from all directions in the $(w\ee ,h\E )$ plane.
When coming from low $w\ee $ values,
the system oscillates and converges to $F$ through region $\cal P$.
When coming from high $w\ee $ values,
the system reaches $F$ through region $\cal T$,
where it bounces back and forth
between the high- and low-activity point attractors
(an oscillation much slower than in $\cal P$).

The nullcline diagram for point $F$ of Figure 6b is illustrated in Figure 6c.
There are now two points of contact between the nullclines,
a situation more degenerate than the one
that obtains from regulating $w\ee $ only,
but `equivalent' to the situation
obtained in the reduced system 
by regulating a single parameter, $w\ee $ or $w\ie $
(compare Figure 6b to Figure 2b).
What characterizes the dynamics at point $F$
is that the system is on the verge of oscillation
and on the boundary of each of the two steady-firing phases.

We finally consider the system
with the four parameters $h\E $, $h\I $, $w\ee $ and $w\ie $ regulated.
We thus include, in addition to equations
\ref{sys_full}, \ref{covee}, \ref{regee}, \ref{covie}, \ref{regie} and \ref{reghe},
a regulation equation for the inhibitory threshold $h\I $:
\begin{equation}
\frac{dh\I (t)}{dt} = \varepsilon\I ( \bar{\sigma}(t) - \theta\I  ).
\label{reghi}
\end{equation}
As in equation \ref{reghe},
the rate constant $\varepsilon\I$ is positive and small,
and $\theta\I $ is chosen in the interval $(.2,.8)$,
with $\theta\I \approx \theta\E $.
The variables now include the activity state $(s,\sigma)$
as well as the four regulated parameters
$h\E $, $h\I $, $w\ee $ and $w\ie $.

Figure 7 illustrates the behavior of this system projected
on the $(w\ee ,w\ie )$ plane,
for the following parameter values:
$w\ei = 10$, $w\ii = 6$, $\rho=.05$, $\theta\ee =.01$, $\varepsilon\ee =.01$,
$\theta\ie =.01$, $\varepsilon\ie =-.005$,
$\theta\E = .5$, $\varepsilon\E = .005$,
$\theta\I = .5$, $\varepsilon\I = .002$.
In the sequel, this parameter setting will be referred to as {\em standard.}
In a first stage, the system converges to a doubly critical point $F$
as described above;
each such point $F$ belongs to the common boundary of the regions
of oscillation, high steady firing, and low steady firing.
Although we cannot thoroughly characterize the surface of $F$ points
in the four-dimensional $(w\ee , w\ie , h\E , h\I )$ space
as we did in the $(w\ee ,w\ie )$ plane for the reduced system,
there is, as remarked above,
a functional equivalence with the $\cal S$ surface.
Note that the projection of the $F$ surface on the $(w\ee ,w\ie )$ plane
has a shape quite similar to that of $\cal S$ in the reduced system.
As before, when the system reaches a point $F$,
all variables settle in a slow, synchronous, almost-periodic motion.
The oscillation of $s$ and $\sigma $ is a nearly rectangular wave,
the system spending nearly all its time in the two corners of the square,
where the relative amount of time spent in each corner
is determined according to the value of parameter $\theta\E (\approx \theta\I )$.
As before too, the first stage,
which consists of the convergence to a doubly critical point $F$,
is robust against parameter changes;
most parameters can be individually
varied over several orders of magnitude
without qualitatively affecting this part of the behavior.

The second stage, consisting of a much slower motion on the $F$ surface,
depends on the values of the various parameters.
For most parameter settings,
including the standard set (see above),
the behavior on this critical surface is a slow, {\em simple,} periodic motion,
of large amplitude in $(s,\sigma)$
and very small amplitude in $(w\ee ,w\ie )$.
The system eventually settles in a periodic attractor of this simple type,
denoted again $G$ in Figure 7.
Figure 8a shows the $(s,\sigma)$ projection of this attractor
for the standard parameter set;
its $(w\ee ,w\ie )$ projection
is a small cycle around point $G$,
whose nullcline diagram
is similar to the one shown in Figure 5 (largely overlapping nullclines).

There exists however a small region of parameter space,
mostly around $\varepsilon\I \approx \varepsilon\E$,
for which a variety of more complex behaviors
are observed during the second stage.
The following two cases are examples of such complex behavior.
For parameters as above (standard) except that
$\varepsilon\E = .0051$,
$\varepsilon\I = .0046$,
and $\theta\ee =.011$,
the system settles in a complex quasi-periodic motion (Figure 8b).
For parameters as standard except that $\varepsilon\I = \varepsilon\E = .005$,
the system displays strongly chaotic behavior (Figure 8, c--e).\footnote
{This behavior takes place only
for {\em some} initial values in the $(w\ee ,w\ie )$ plane;
other initial values converge to a point attractor.}
Both of these behaviors are actually {\em attractors,}
reached after considerable time,
yet similar behaviors also take place
while the system is still moving slowly on the critical surface.

To summarize, both in the reduced and in the full system,
convergence to a doubly critical surface
between the regions of fast oscillations
and of high and low steady firing
takes place reliably for a broad range of parameters.
Once this doubly critical surface is reached,
motion becomes slow,
depends on parameters,
and, when examined in detail,
reveals a variety of behaviors,
ranging from simple periodic firing to chaos.

\section{Discussion}
\label{disc}

This paper proposes that a regulation mechanism
underlies criticality in brain dynamics.
In such a scheme, regulation stabilizes the dynamics near an instability.
The force driving the system towards criticality
is a covariance-governed modification of synaptic efficacies
in a recurrent network.
Although it has been argued that criticality
in some physical systems may be self-organized (Bak et al. 1987),
this phenomenon may not be very widespread.
The nervous system is moreover regulated homeostatically
to withstand perturbations of various sorts.
It is then of interest to explain
how criticality in brain dynamics may nevertheless arise,
from a specific, well-documented, mechanism of synaptic plasticity.

The chief motivation for viewing the dynamics of the nervous sytem as critical
is the observation that brains are very sensitive organs.
Not only do our brains draw distinctions between stimuli
that differ only in minute details,
they also allow us to establish subtle yet clear-cut boundaries
between cognitive categories at different levels.
The various manifestations of hyperacuity in sensory systems
may be no more than elaborate forms of signal amplification;
however, in higher cognitive functions
such as language, abstract thinking, or, for instance, artistic composition,
the ability to create a new category by drawing a fine line---an
ability which manifests itself early in life
and stays with us for a long time---argues
in favor of regulated criticality.
Such a mechanism appears to be necessary
in order to explain how sensitivity is {\em maintained}
in the face of the profound changes
that affect the connectivity of the brain
throughout development and learning.
If no such mechanism were present,
one would expect that the ongoing modification
of the networks which carry mental representations
would soon bring these networks to generic states;
as mentioned in Section \ref{intro},
a dynamical system in a generic state does not show high susceptibility
to external influences.

The emergence of new cognitive categories
may in effect be likened to a process of morphogenesis in embryology,
or differentiation in cell biology.
A biological structure that is about to undergo differentiation
is at that particular instant of time unstable,
and, as the well-studied mechanism of {\em induction} shows,
highly susceptible to external signals.
From a dynamical-system perspective,
the emergence of qualitatively new behavior,
e.g., the splitting of one attractor into two, is a bifurcation.
The complexification of an individual's cognitive apparatus
in the course of his or her life
may be viewed as an open-ended sequence of such bifurcations.
Such an interpretation has been defended by Ren\'e Thom (1975),
and related ideas have been expressed by several authors
(e.g. van der Maas and Molenaar 1992).
Thom (1975) also suggested that structurally stable non-generic singularities
may arise from a process he termed the {\em stabilization of thresholds;}
this process itself would result, in various biological contexts,
from the reinforcment of homeostatic mechanisms.\footnote
{We thank Jean Petitot for pointing out to us
that regulated criticality as proposed here
is closely related to Thom's ideas.}

The covariance plasticity rule we use is linear and straightforward.
Equation \ref{regee} may be viewed
as a mean-field version of the covariance rule
used in the associative-memory literature
(see e.g. Willshaw and Dayan 1990).
However, we make a rather different use of this rule.
In an associative-memory model,
pre- and post-synaptic activities are generally assumed to be independent,
yielding a zero expected value for the covariance.
Weights are modified according to the {\em instantaneous} covariance,
and, as noted in Dayan and Sejnowski (1993),
storage is marked by the departures of the empirical average of this quantity
from its expected value, which is zero.
In our model,
the expected covariance is positive in the oscillatory phase.
The regulation mechanism acts on a {\em slow} time scale,
and, although we use the instantaneous covariance
in the modification rule (Equation \ref{regee}),
we might as well have used the time-averaged covariance;
fast variations of the instantaneous covariance are actually smoothed out
in the integration of the differential equation.
Of course by the very principle of regulation proposed,
the system does not dwell in the oscillatory phase;
in the regulated state, the average covariance is low.

The other major difference between the situation studied here
and the associative-memory paradigm
is the assumption of uniform weights.
As noted in Section \ref{regulation},
the covariance in our uniform-weight network
is simply the variance of the population-averaged activity about its mean,
and is always non-negative.
This makes it necessary to subtract from it a positive constant $\theta\ee$
in order to allow for {\em decreases} of the weights.
Thus, whereas in associative-memory models a synaptic weight
decreases as a result of {\em negative} instantaneous covariance
between the pre- and post-synaptic neurons,
the condition for weight decrease in our model
is that the mean covariance be small or zero,
which happens when the system is at rest in a point attractor,
of either low or high activity.

The uniform-weight network used in the present study
lends itself to a detailed mathematical/numerical analysis.
We performed a bifurcation analysis of the continuous-time differential system
that describes the behavior of this network in the thermodynamic limit.
This analysis (Section \ref{full}) reveals, among other features,
the existence of a critical surface ${\cal S}_h$ in parameter space,
where the system undergoes an abrupt transition
from oscillatory behavior to high-rate steady firing.
We showed (Section \ref{full})
that Hebbian modification of the E-to-E synaptic weights
drives the system toward this surface ${\cal S}_h$;
this is the main mechanism of regulated criticality proposed.

However, when the system is at a general position on ${\cal S}_h$,
it spends most of its time in the high-activity state,
which is undesirable.
A more realistic situation results from introducing some form of symmetry
between the high- and low-activity phases,
making use of the firing thresholds.
We investigated two ways---formally
different but functionally equivalent---to do this.
The mathematically simpler way
is to enforce an accurate symmetry on the dynamics,
by imposing an appropriate relationship between
the firing thresholds and the synaptic weights (equations \ref{chvar}).
This results in a {\em reduced} system, with only four parameters;
in this system,
there occurs a {\em double} bifurcation when the system traverses
the critical surface---now denoted $\cal S$---separating
the oscillatory phase from the bistable, high/low, steady-firing phase
(Section \ref{model}).
Regulation of the sole E-to-E weight brings the system
to this doubly critical surface $\cal S$ (Section \ref{reduced}).

A biologically more satisfactory solution
is to {\em regulate} one or both of the firing thresholds
so as to control the mean firing rates (Section \ref{full}).
Thus, when we regulate the threshold for the excitatory neurons
in addition to the E-to-E weight,
the system converges to the intersection of ${\cal S}_h$
with another critical surface, ${\cal S}_l$,
which separates the oscillatory phase from the low-activity fixed-point region.
Intersection points between ${\cal S}_h$ and ${\cal S}_l$
are again doubly critical,
and they attract the system for a wide range of parameter values.

When the system is on this doubly critical surface,
it takes only a small weight perturbation
to induce either of the three behaviors:
intrinsic oscillation (region $\cal P$ of section \ref{full}),
high activity, quiescence.
It is easily seen that, when in this state,
the network can also be efficiently driven
by a small-amplitude time-varying signal, i.e., an external field;
it is thus highly sensitive to input.
The stabilization at the boundary of a region of oscillatory behavior
appears to be consistent with at least one conclusion
that can be safely drawn from the recent literature on cortical oscillations,
namely the fact that the precise conditions under which
these oscillations occur are difficult to pin down.

We further investigated the effect of regulating
the E-to-I weight in addition to the E-to-E weight,
according to a similar covariance rule.
We showed that regulating these two weights
as well as the two firing thresholds results,
under appropriate parametric conditions,
in convergence to an even more degenerate state.
When the system is in that state,
its flow vanishes on an entire one-dimensional curve
in the two-dimensional phase space,
instead of on isolated points.
This convergence is slow and parameter-dependent,
yet it is interesting to note that
when the system is in or near this highly degenerate state
it exhibits a range of diverse behaviors,
including chaos (Section \ref{full}).
The chaotic behavior shown in Figure 8, c--e
consists of an irregular sequence of spontaneous transitions
between the three fundamental phases of the system:
oscillatory, high-activity, low-activity.

While the uniform-weight network studied in this paper
lends itself to a convenient mathematical analysis,
it would be interesting to know whether critical behavior
may arise from {\em local} covariance plasticity,
where synaptic changes are made to depend on pre- and post-synaptic activities
relative to individual synapses.
This question should be focused by considerations
about the elaborate forms of input sensitivity
that could play a role in higher brain functions.

{\bf Acknowledgments} Important contributions in the early stages of this work
were made by Howard Gutowitz.
We thank Rob de Boer for making available to us the use of the software {\em GRIND,}
which proved an extremely valuable tool in this study.
Various suggestions were made by a number of colleagues,
from whom we wish to thank in particular G\'erard Weisbuch, Jean Petitot,
G\'erard Toulouse, Jean-Pierre Nadal, Claude Meunier, David Hansel,
Ha\"{\i}m Sompolinsky, and Stuart Geman.

\vglue 1cm

\noindent{\bf References}

\begin{description}

\item Artola, A., Br\"ocher, S., and Singer, W. 1990.
Different voltage-dependent thresholds for inducing long-term depression
and long-term potentiation in slices of rat visual cortex.
{\em Nature}, {\bf 347}, 69--72.

\item Bak, P., Tang, C., and Wiesenfeld, K. 1987.
Self-Organized Criticality: An Explanation of $1/f$ Noise.
{\em Phys. Rev. Lett.}, {\bf 59}, 381--384.

\item Bienenstock, E., Cooper, L.N., and Munro, P. 1982.
Theory for the development of neuron selectivity:
Orientation specificity and binocular interaction in visual cortex.
{\em J. Neurosci.}, {\bf 2}, 32--48.

\item Borisyuk, R.M., and Kirillov, A.B. 1992.
Bifurcation analysis of a neural network model.
{\em Biol. Cybernetics} {\bf 66}, 319--325.

\item Dayan, P., and Willsahw, D. 1991.
Optimising synaptic learning rules in linear associative memories.
{\em Biol. Cybernetics} {\bf 65}, 253--265.

\item Dayan, P., and Sejnowski, T.J. 1993.
The Variance of Covariance Rules for Associative Matrix Memories
and Reinforcement Learning.
{\em Neural Computation}, {\bf 5}, 205--209.

\item Dudek, S.M., and Bear, M.F. 1992.
Homosynaptic long-term depression in area CA1 of hippocampus
and effects of {\em N}-methyl-D-aspartate receptor blockade.
{\em Proc. Natl. Acad. Sci. USA}, {\bf 89}, 4363--4367.

\item Glauber, R.J. 1963.
Time-dependent Statistics of the Ising Model.
{\em J. Math. Phys.} 4, 294--307.

\item Hebb D.O. 1949.
{\em The Organization of Behavior.}
Wiley, New York.

\item Fr\'egnac, Y., Shulz, D., Thorpe, S., and Bienenstock, E. 1988.
A cellular analogue of visual cortical plasticity.
{\em Nature}, {\bf 33}, 367--370.

\item Fr\'egnac, Y., Shulz, D., Thorpe, S., and Bienenstock, E. 1992.
Cellular analogs of visual cortical epigenesis.
I--Plasticity of orientation selectivity.
{\em Journal of Neuroscience}, {\bf 12}, 1280--1300.

\item Guckenheimer, J., and Holmes, P. 1983
{\em Nonlinear oscillations, dynamical systems,
and bifurcations of vector fields.}
Springer-Verlag, New York.

\item Langton, C.R. 1990.
Computation at the edge of chaos:
Phase transitions and emergent computation.
{\em Physica D}, {\bf 42}, 12--37.

\item Linsker, R. 1986.
From basic network principles to neural architecture:
Emergence of spatial opponent cells.
{\em Proc. Natl. Acad. Sci. USA}, {\bf 83}, 7508--7512.

\item Metzger, Y., and Lehmann, D. 1990.
Learning Temporal Sequences by Local Synaptic Changes.
{\em Network: Computation in Neural Systems}, {\bf 1}(3), 169--188.

\item Metzger, Y., and Lehmann, D. 1994.
Learning Temporal Sequences by Excitatory Synaptic Changes Only.
{\em Network: Computation in Neural Systems}, {\bf 5}, 89--99.

\item Rubin, N. 1988.
Equilibrium and oscillations in stochastic neural networks.
Master's thesis, The Hebrew University, Jerusalem, Israel (in Hebrew).

\item Schuster, H.G., and Wagner, P. 1990
A model for neuronal oscillations in the visual cortex.
1. Mean-field theory and derivation of the phase equations.
{\em Biological Cybernetics}, {\bf 64}, 77--82.

\item Sejnowski, T.J. 1977a.
Storing covariance with nonlinearly interacting neurons.
{\em J. Math. Biol.}, {\bf 4}, 303--321.

\item Sejnowski, T.J. 1977b.
Statistical constraints on synaptic plasticity.
{\em J. Theor. Biol.}, {\bf 69}, 385--389.

\item Stanton, P.K. and Sejnowski, T.J. 1989.
Associative long-term depression in the hippocampus
induced by hebbian covariance.
{\em Nature}, {\bf 339}, 215, May 1989.

\item Thom, R. 1975.
{\em Structural Stability and Morphogenesis:
an Outline of a General Theory of Models}.
W. A. Benjamin, Reading, Mass.

\item van der Maas, H.L.J., and Molenaar, P.C.M. 1992.
Stagewise Cognitive Development: An Application of Catastrophe Theory.
{\em Psychological Review}, {\bf 99}(3), 395--417.

\item Willshaw, D., and Dayan, P. 1990.
Optimal Plasticity from Matrix Memories: What Goes Up Must Come Down.
{\em Neural Computation}, {\bf 2}, 85--93.

\end{description}

\renewcommand{\baselinestretch}{1}
\small

\newpage

\centerline{\psfig{file=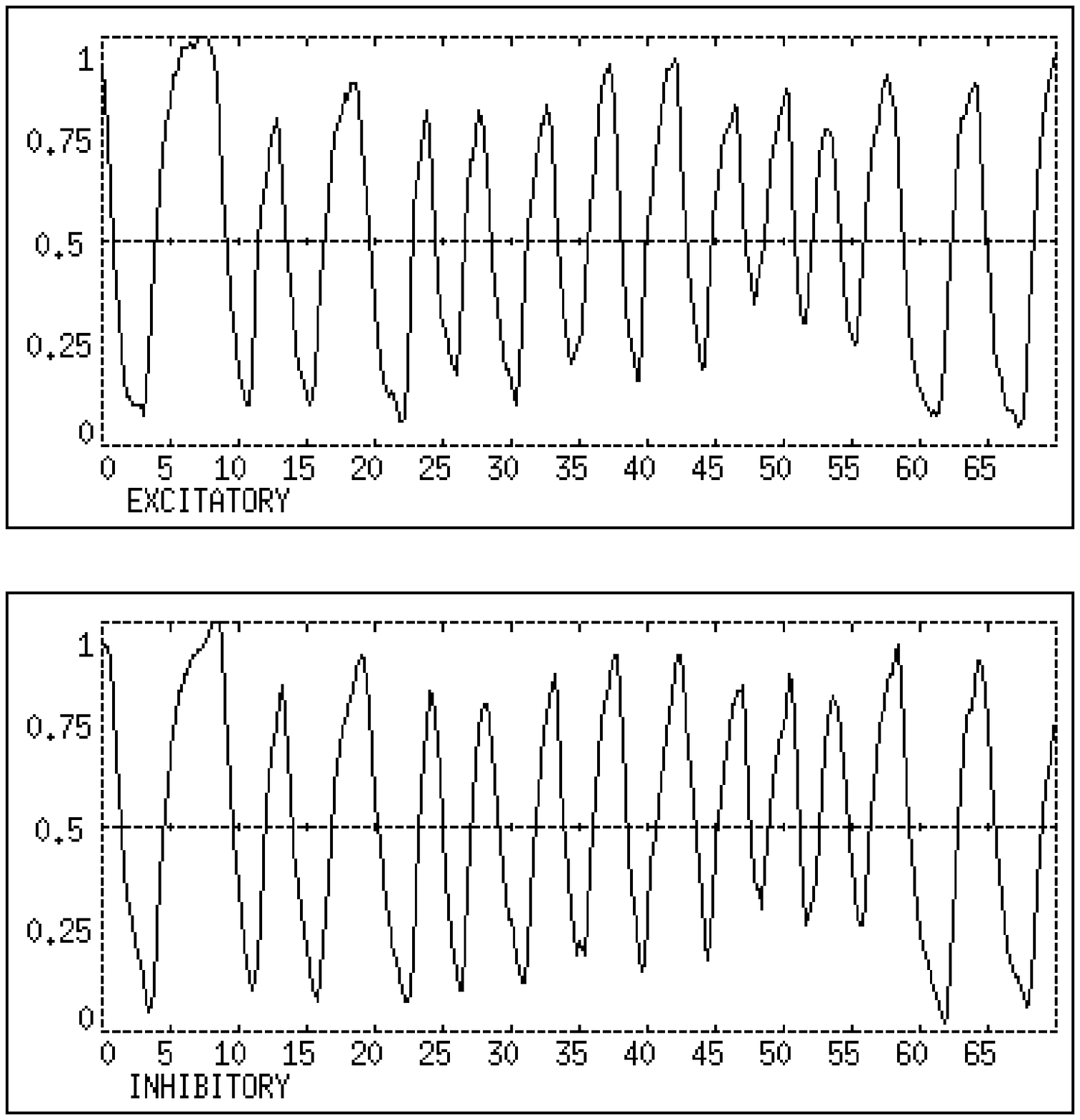,height=20cm}} 
\begin{quote}
Figure 1: Mean activities of excitatory and inhibitory populations
in a moderate-size uniform-weight system
exhibiting oscillatory behavior ($N=70$; Glauber dynamics).
\end{quote}

\newpage

\centerline{\psfig{file=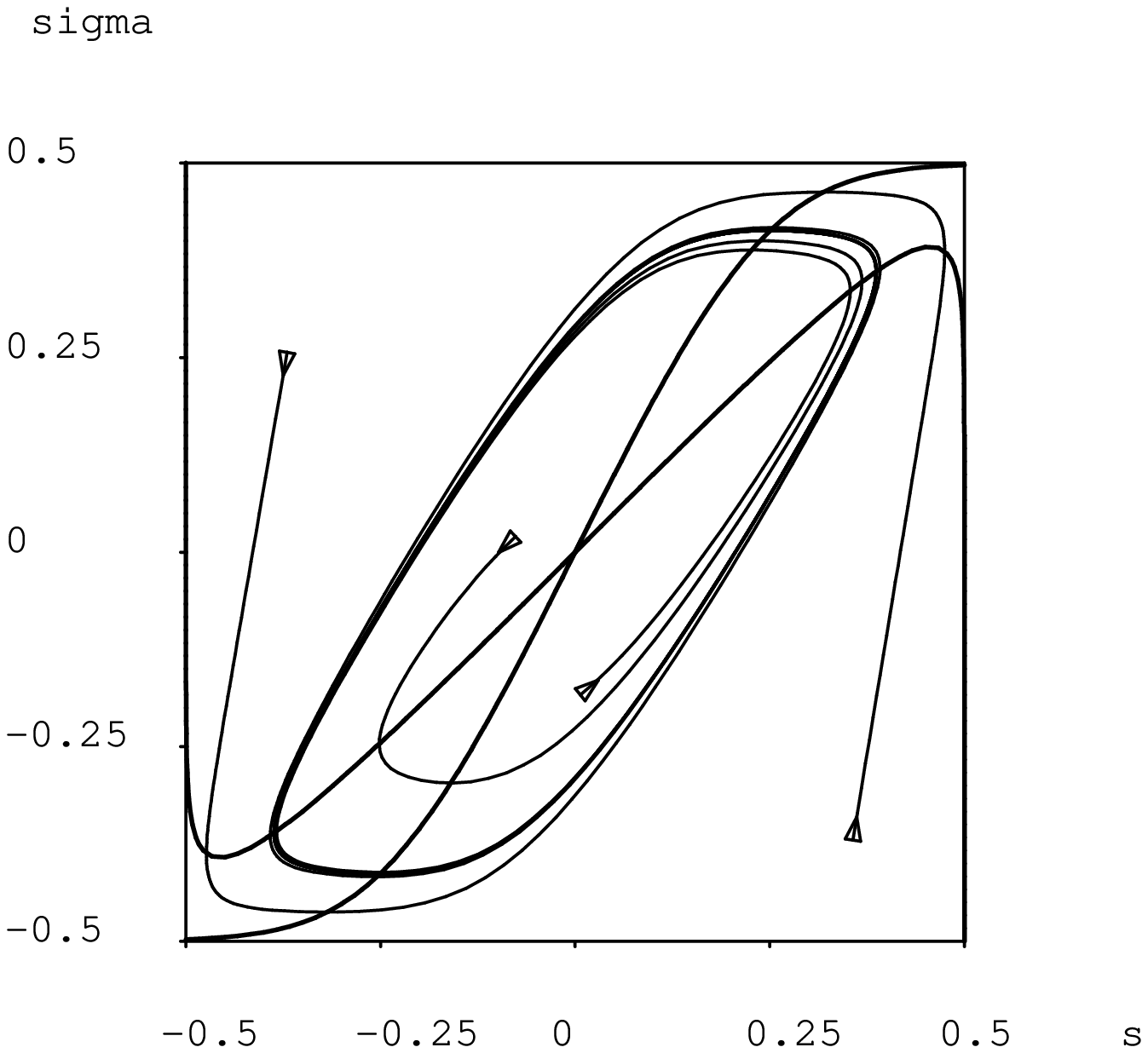,height=6cm}} 
\centerline{\psfig{file=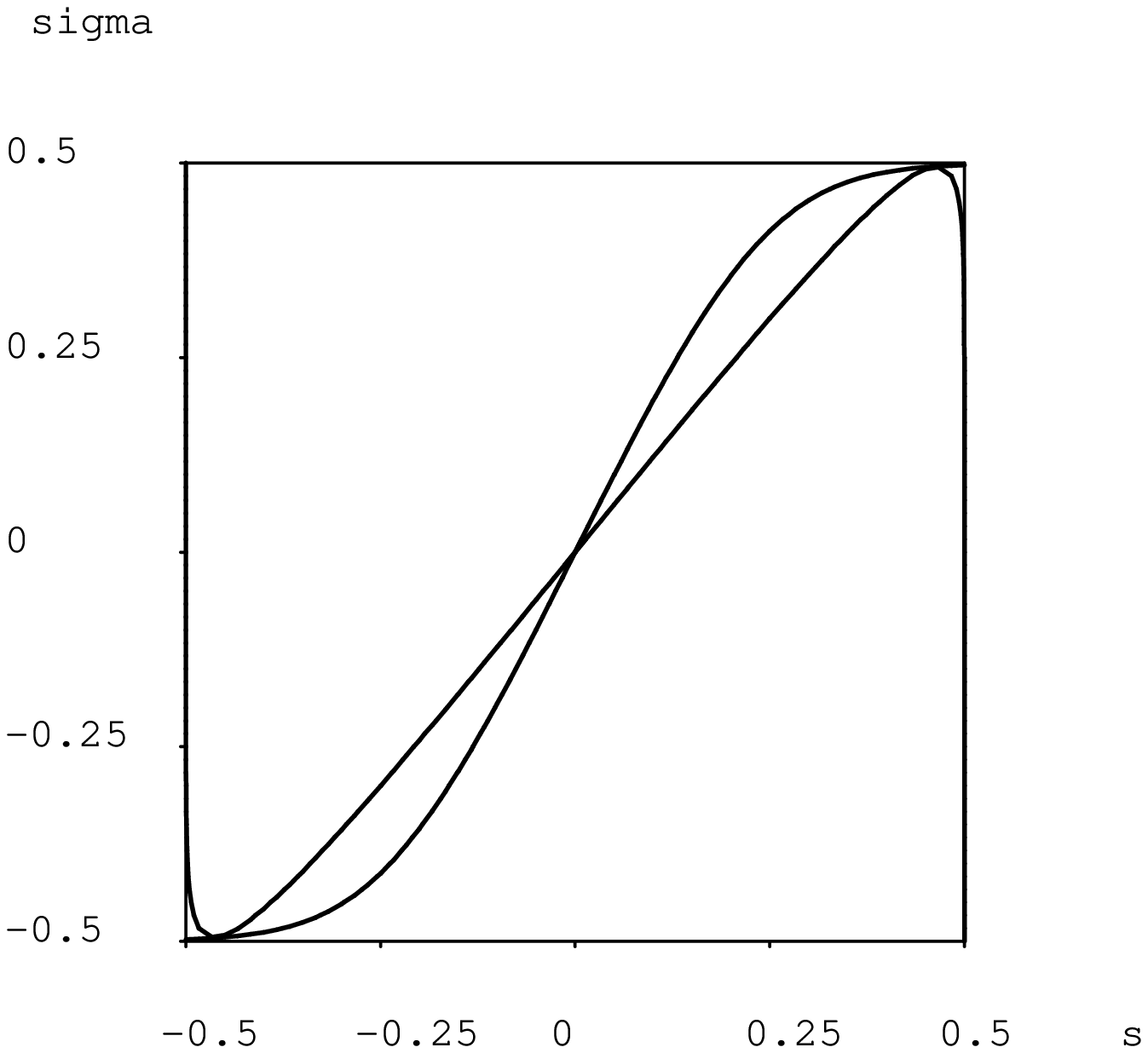,height=6cm}} 
\centerline{\psfig{file=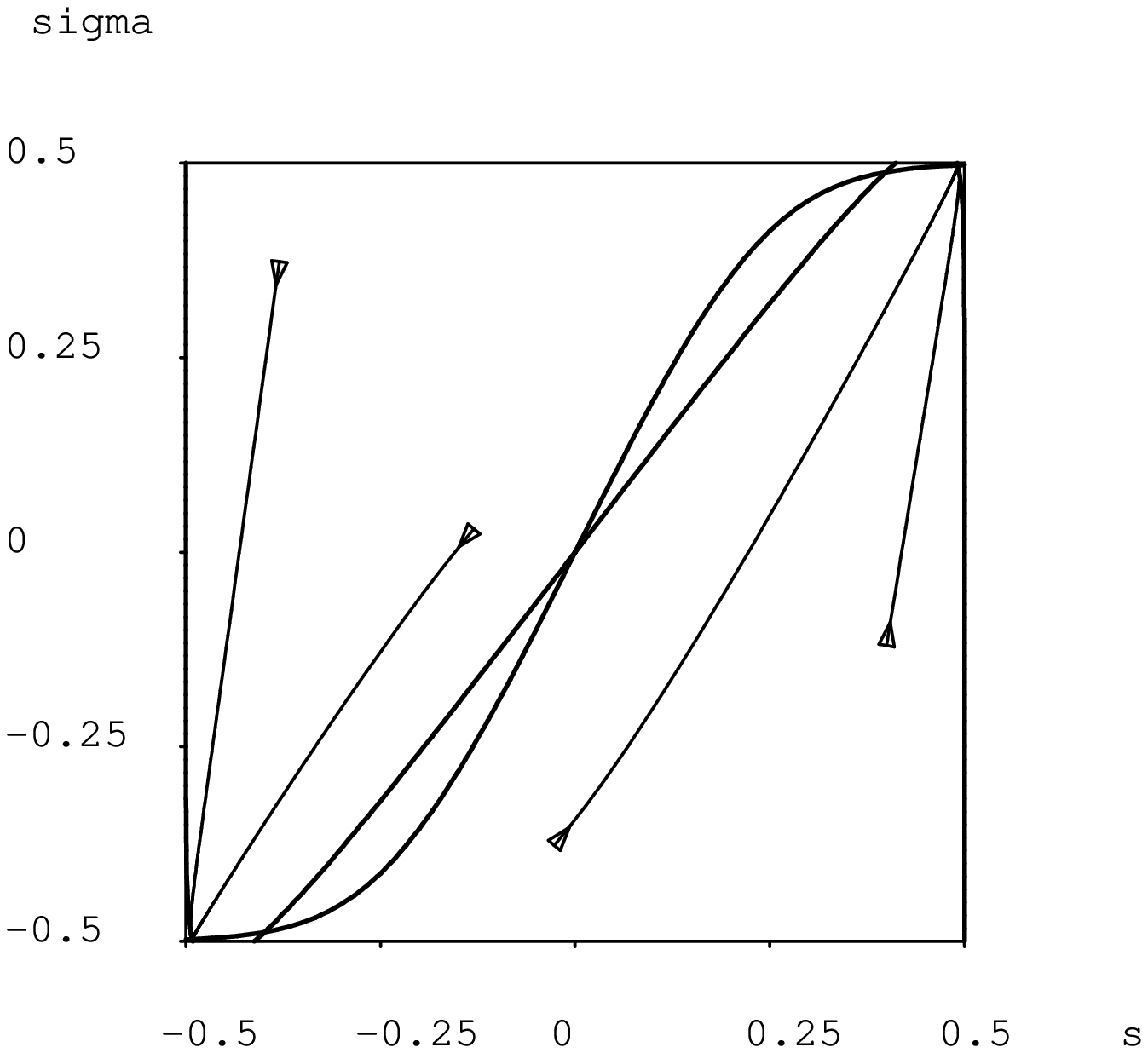,height=6cm}} 
\begin{quote}
Figure 2: Behavior of mean-field system
for different values of the E-to-E synaptic weight $w\ee $.
Diagrams show trajectories and nullclines.
(a) $w\ee = 12$ (parameters are as in Figure 1);
all trajectories converge to a limit cycle.
(b) $w\ee = \hat w\ee \sn \approx 14.22$;
the system is at the saddlenode bifurcation:
nullclines are tangent to each other
(no trajectories shown).
(c) $w\ee = 15$; nullclines intersect,
the periodic attractor has vanished,
two point attractors have appeared.
\end{quote}

\newpage

\centerline{\psfig{file=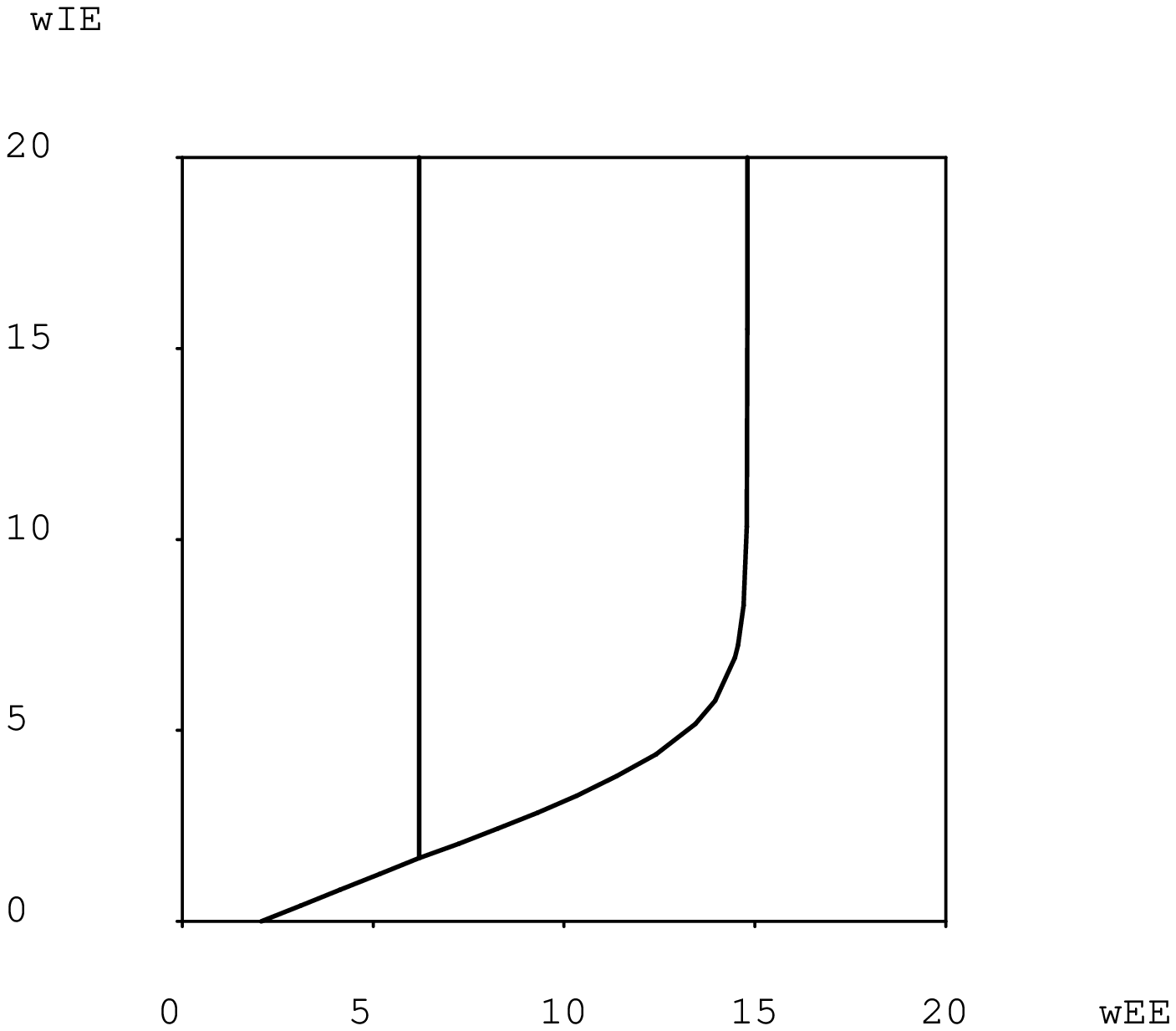,height=8cm}}
\centerline{\psfig{file=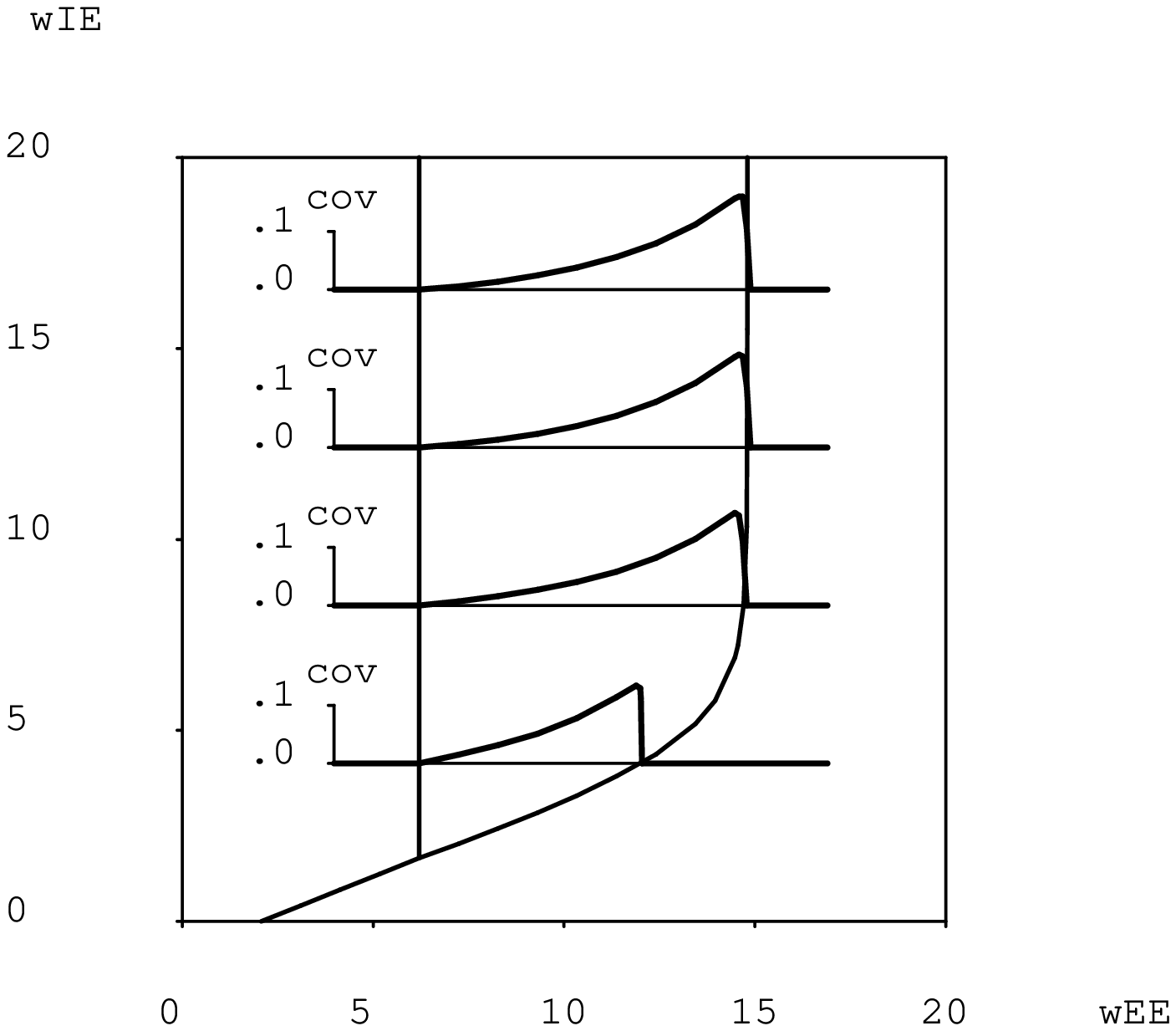,height=8cm}}
\vglue 2cm
\begin{quote}
Figure 3: Bifurcation diagram in $(w\ee ,w\ie )$ plane.
(a) Diagram shows three regions,
characterizing different attractor configurations.
Region $\cal O$: single point attractor, of intermediate activity level;
region $\cal P$: periodic attractor, as depicted in Figure 2a;
region $\cal T$: two point attractors,
of high and low activity, as depicted in Figure 2c.
Transitions between regions occur through bifurcations,
of Hopf type between $\cal O$ and $\cal P$,
of saddlenode type between $\cal P$ and $\cal T$ (curve $\cal S$),
and of pitchfork type between $\cal O$ and $\cal T$.
(b) Average covariance along four different lines of constant $w\ie $
in the $(w\ee ,w\ie )$ plane.
Note the sharp variation of the covariance on the critical line $\cal S$
separating $\cal P$ from $\cal T$.
\end{quote}

\newpage

\centerline{\psfig{file=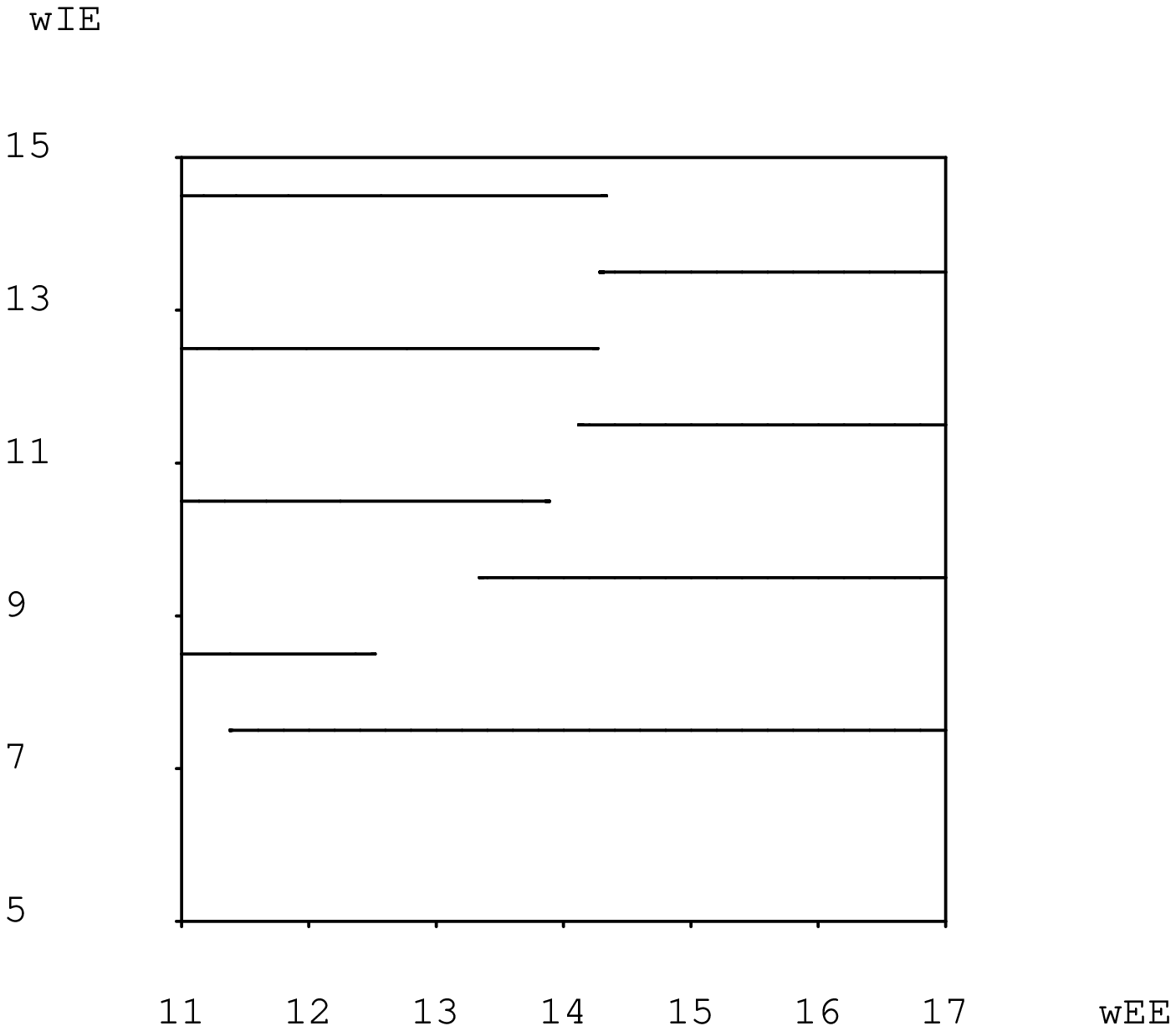,height=6cm}} 
\centerline{\psfig{file=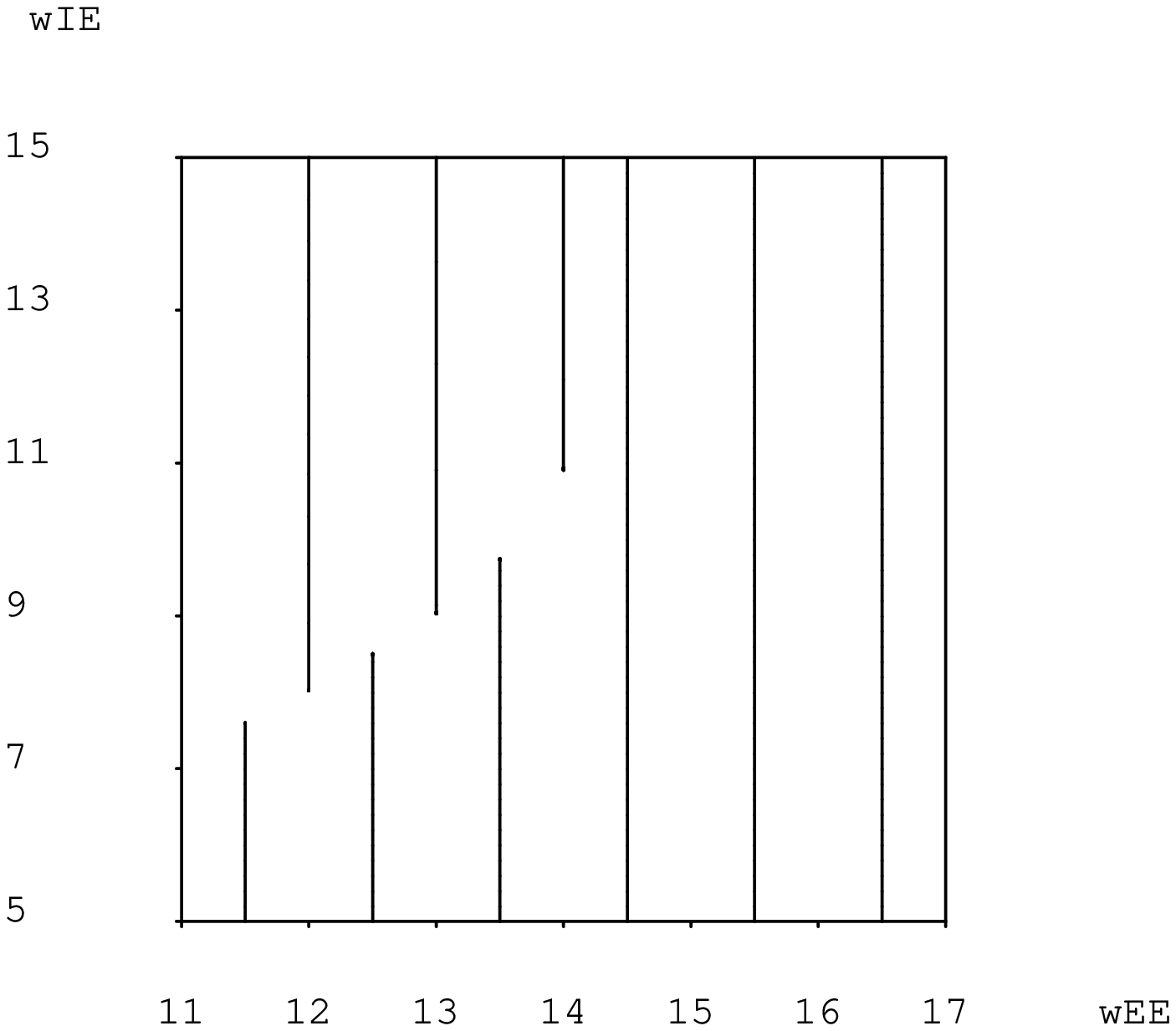,height=6cm}} 
\centerline{\psfig{file=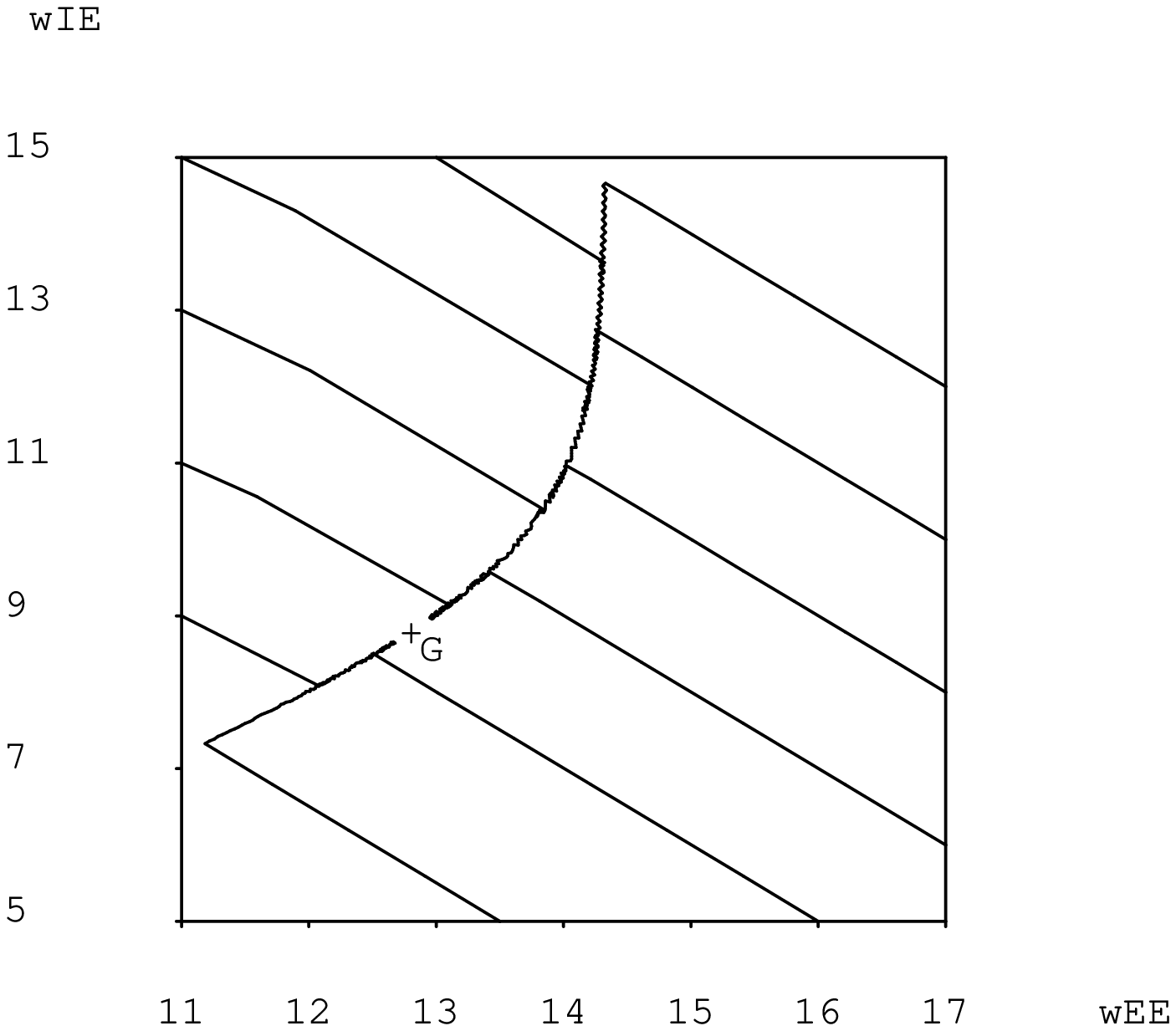,height=6cm}} 
\begin{quote}
Figure 4: Regulation of system \ref{sys_red} by covariance plasticity.
(a) $w\ee $ is regulated, $w\ie $ is constant:
state converges to critical surface $\cal S$.
(b) $w\ie $ is regulated, $w\ee $ is constant:
state converges to critical surface $\cal S$.
(c) both $w\ee $ and $w\ie $ are regulated:
state converges to a particular point, denoted $G$,
on critical surface $\cal S$.
\end{quote}

\newpage \vglue 2cm

\centerline{\psfig{file=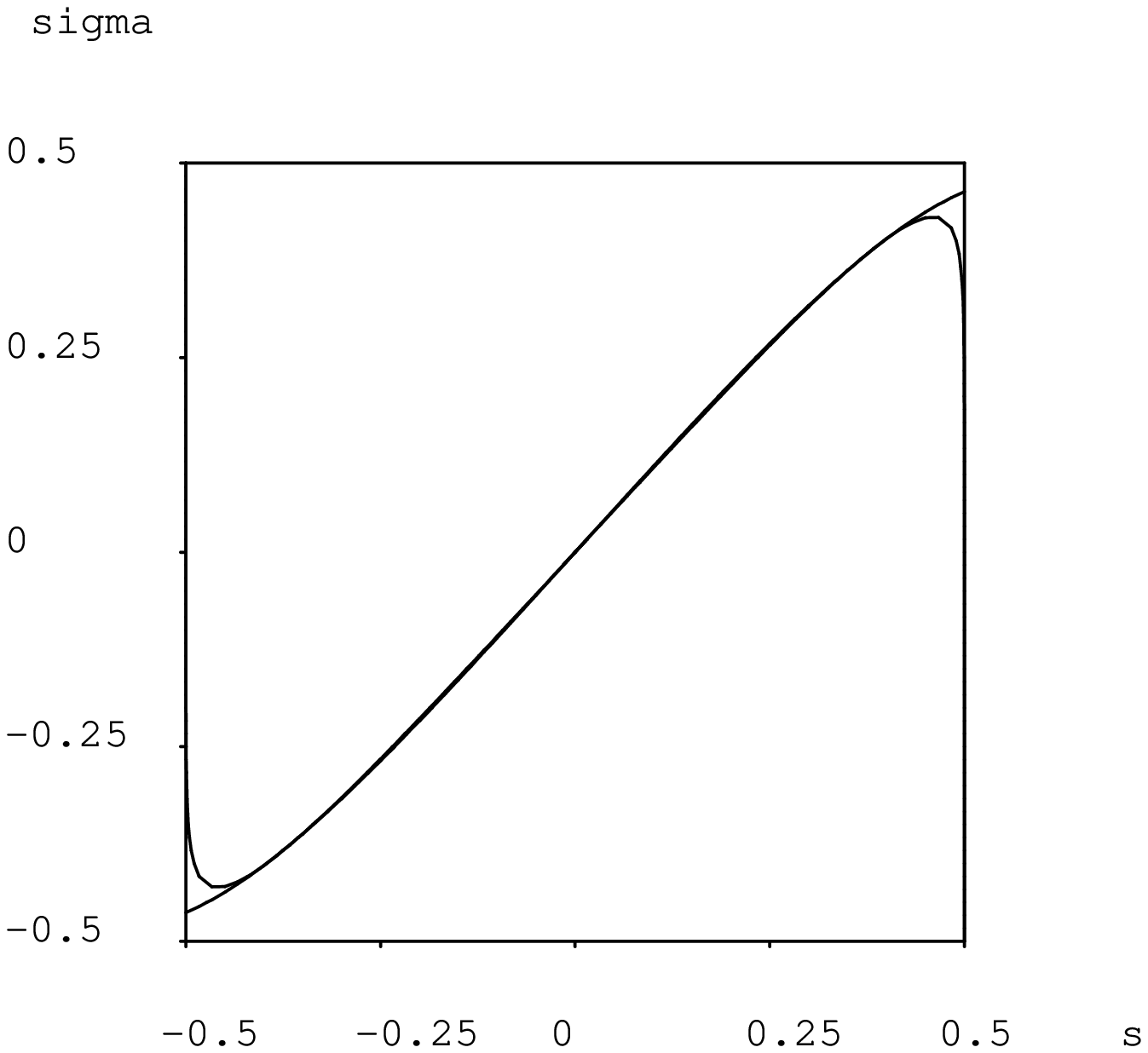,height=16cm}} 
\vglue 2cm
\begin{quote}
Figure 5: Nullcline diagram at point $G$ (see figure 4c).
Nullclines overlap almost perfectly over much of the interval $[-.5,.5]$.
\end{quote}

\newpage

\centerline{\psfig{file=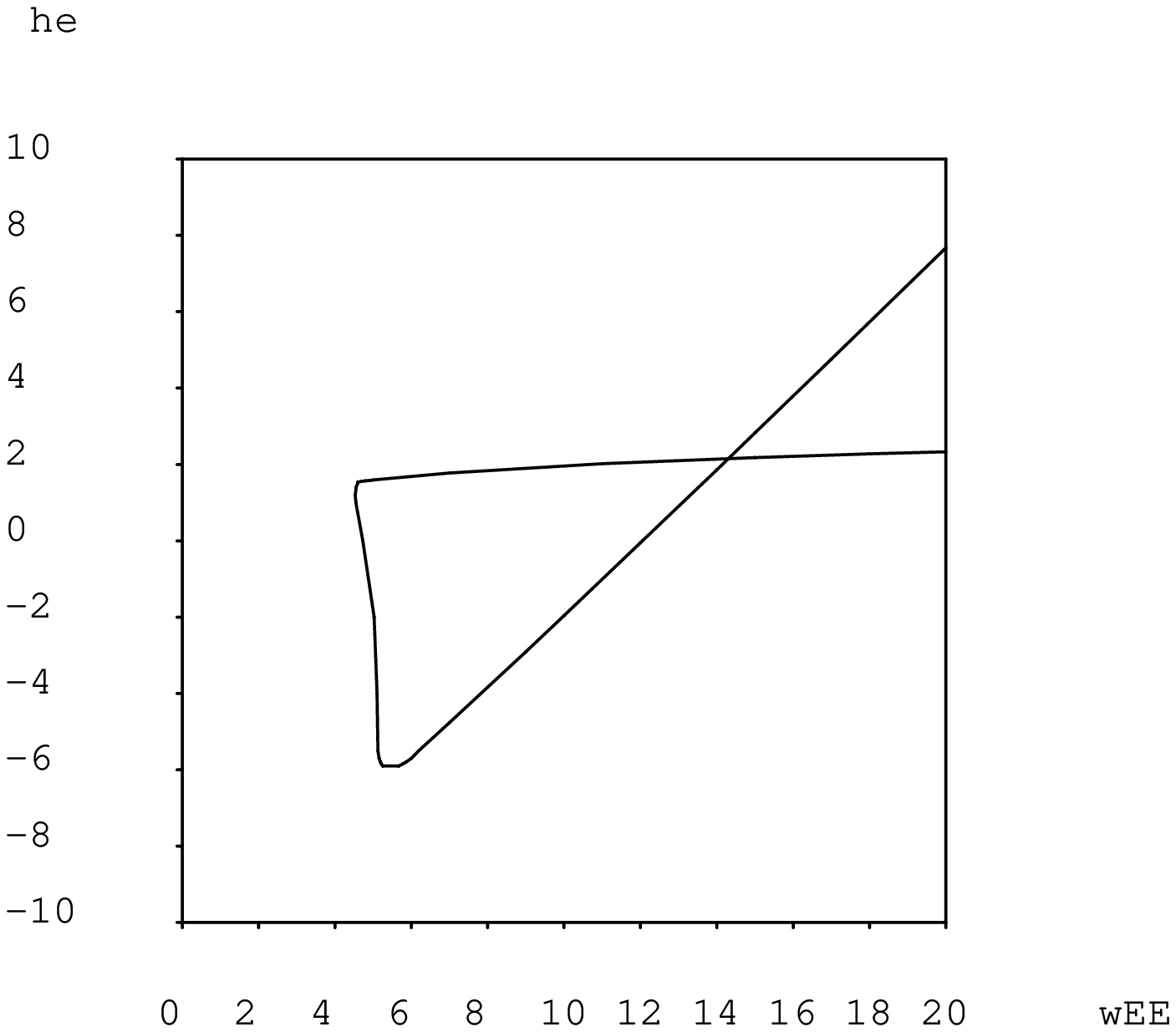,height=6cm}} 
\centerline{\psfig{file=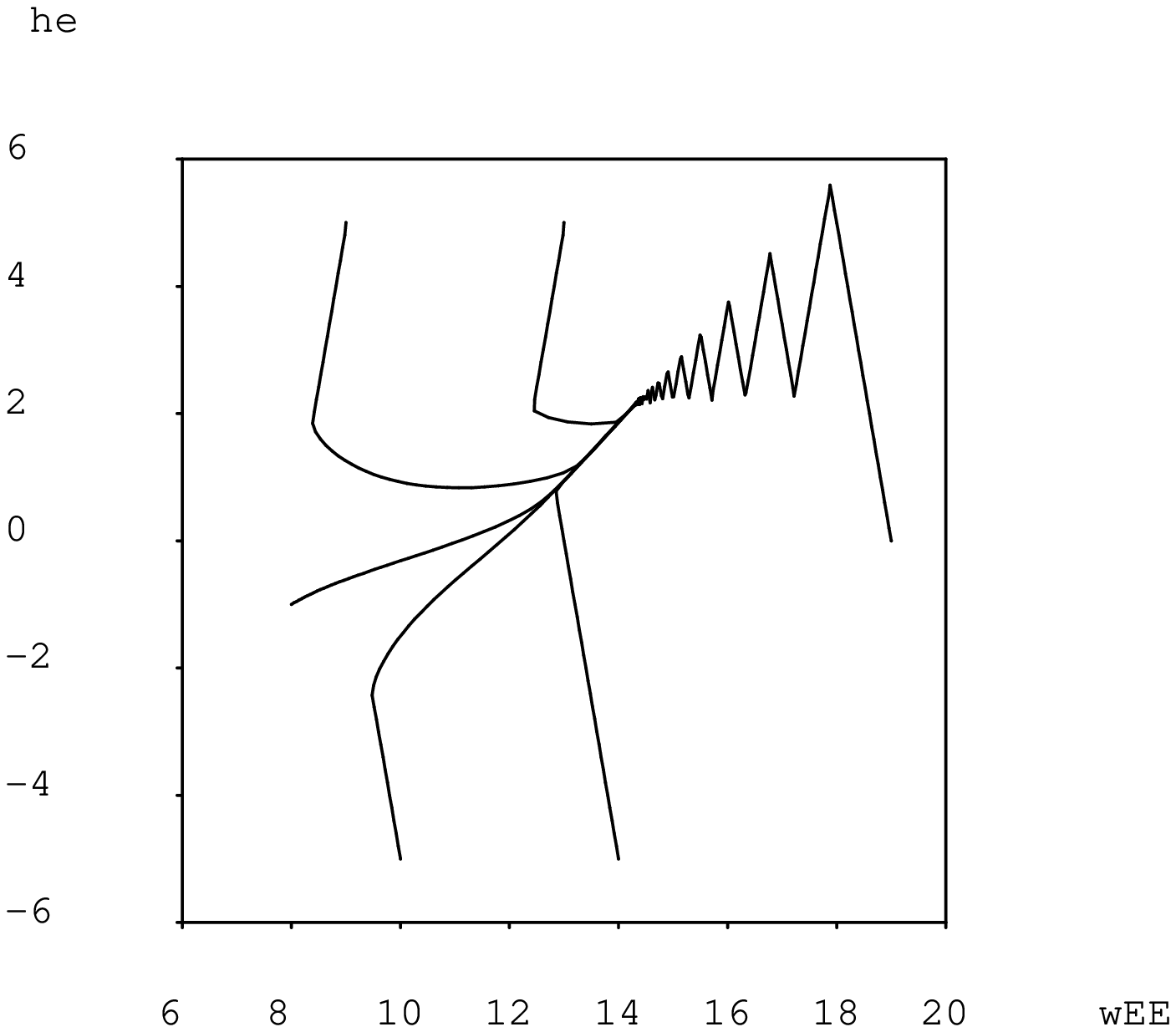,height=6cm}} 
\centerline{\psfig{file=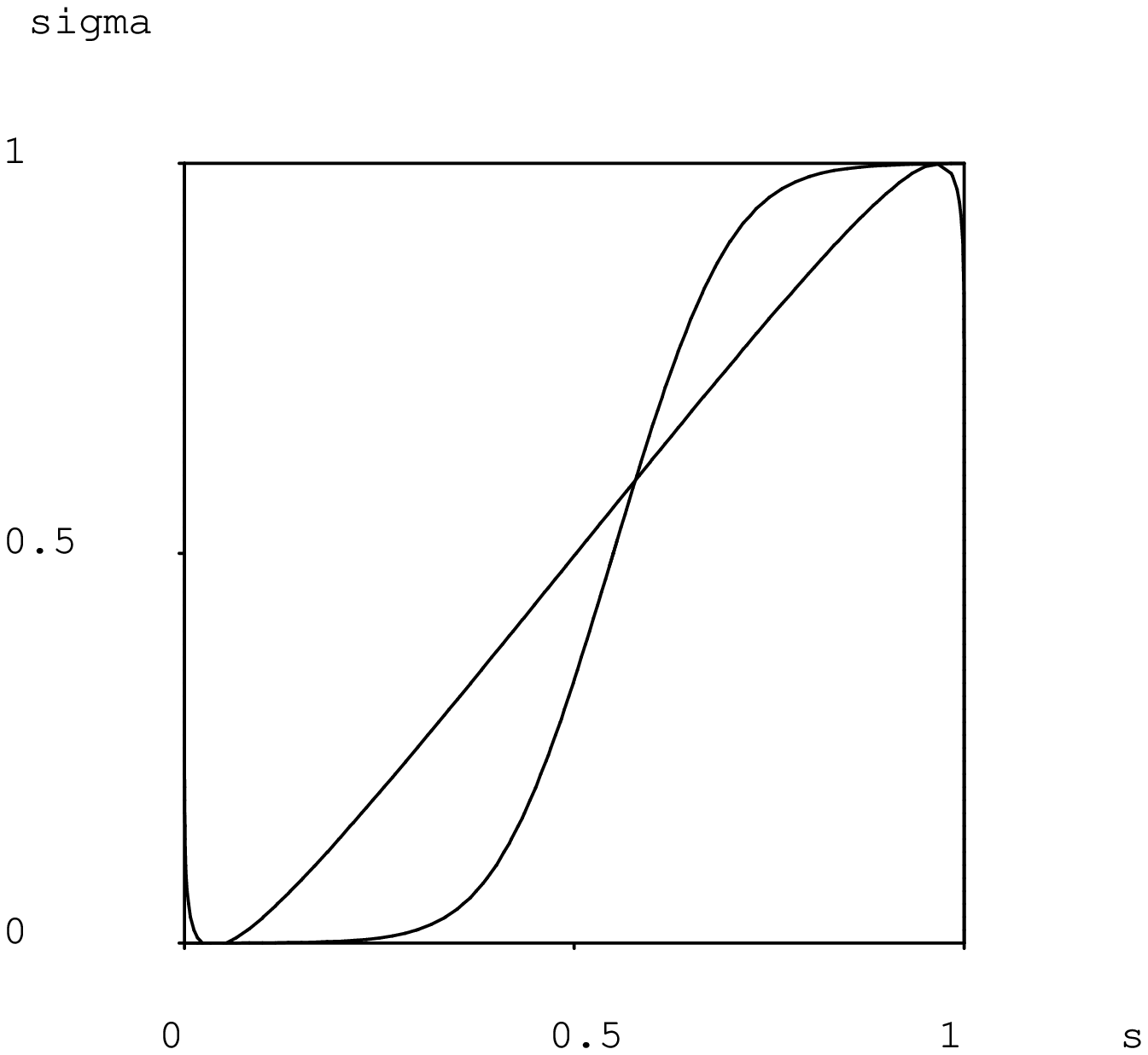,height=6cm}} 
\begin{quote}
Figure 6: Regulation of two parameters in system \ref{sys_full}.
(a) Bifurcation diagram in $(w\ee ,h\E )$ plane.
(b) Regulation of $w\ee $ and $h\E $ causes convergence to point $F$,
the intersection of critical lines ${\cal S}_h$ and ${\cal S}_l$.
(c) Nullcline diagram at $F$.
\end{quote}

\newpage \vglue 2cm

\centerline{\psfig{file=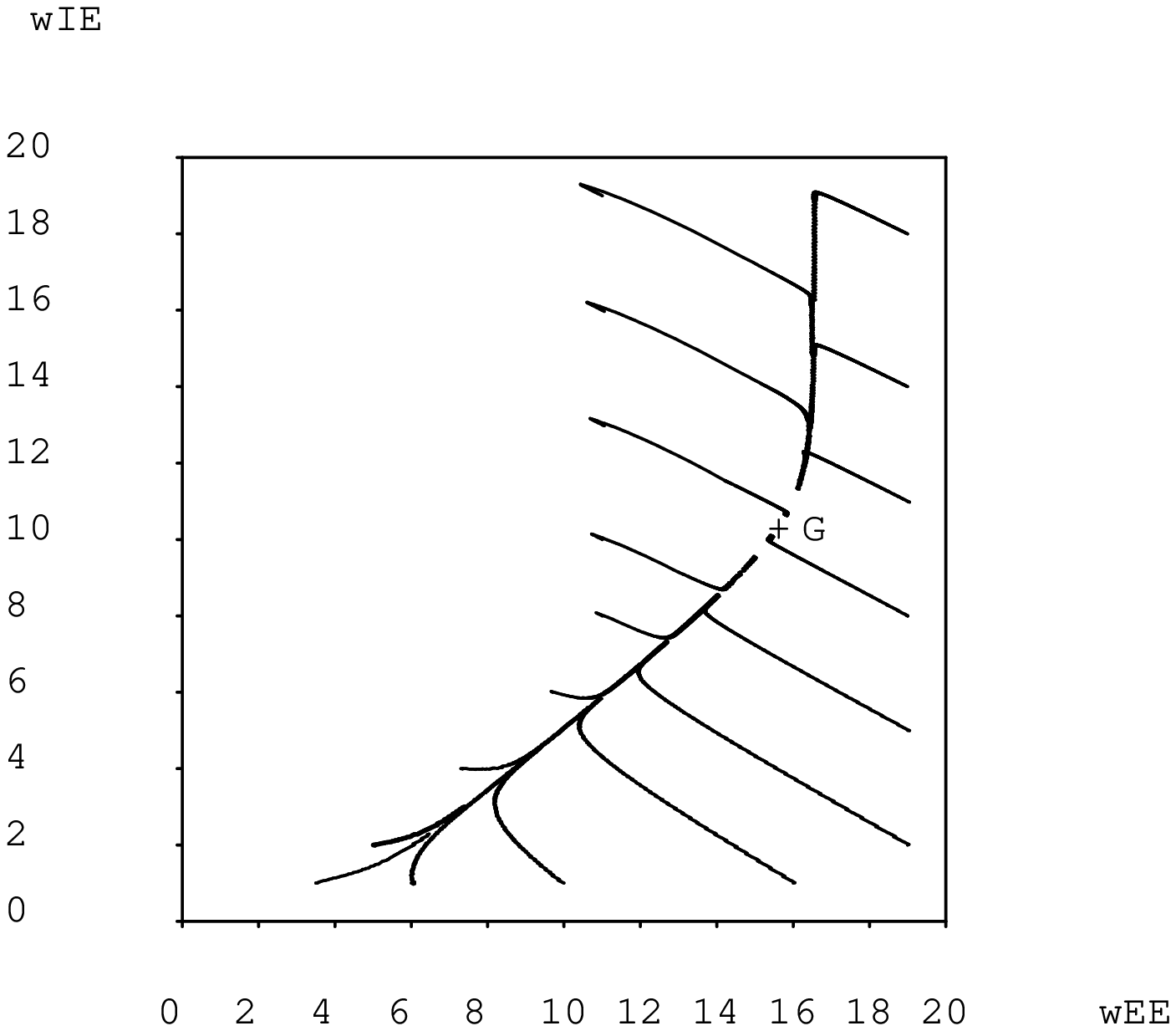,height=16cm}} 
\begin{quote}
Figure 7: Behavior of full system under simultaneous regulation of four parameters.
Diagram shows projection on $(w\ee ,w\ie )$ plane,
illustrating the similarity of behavior with reduced system
(compare with Figure 4c, but note difference of scales).
Limits of the attraction basin to the left are roughly indicated
by the starting points of the trajectories shown;
attraction basin is unbounded in all other directions.
\end{quote}

\newpage

  \begin{tabular}{cc}
    \psfig{file=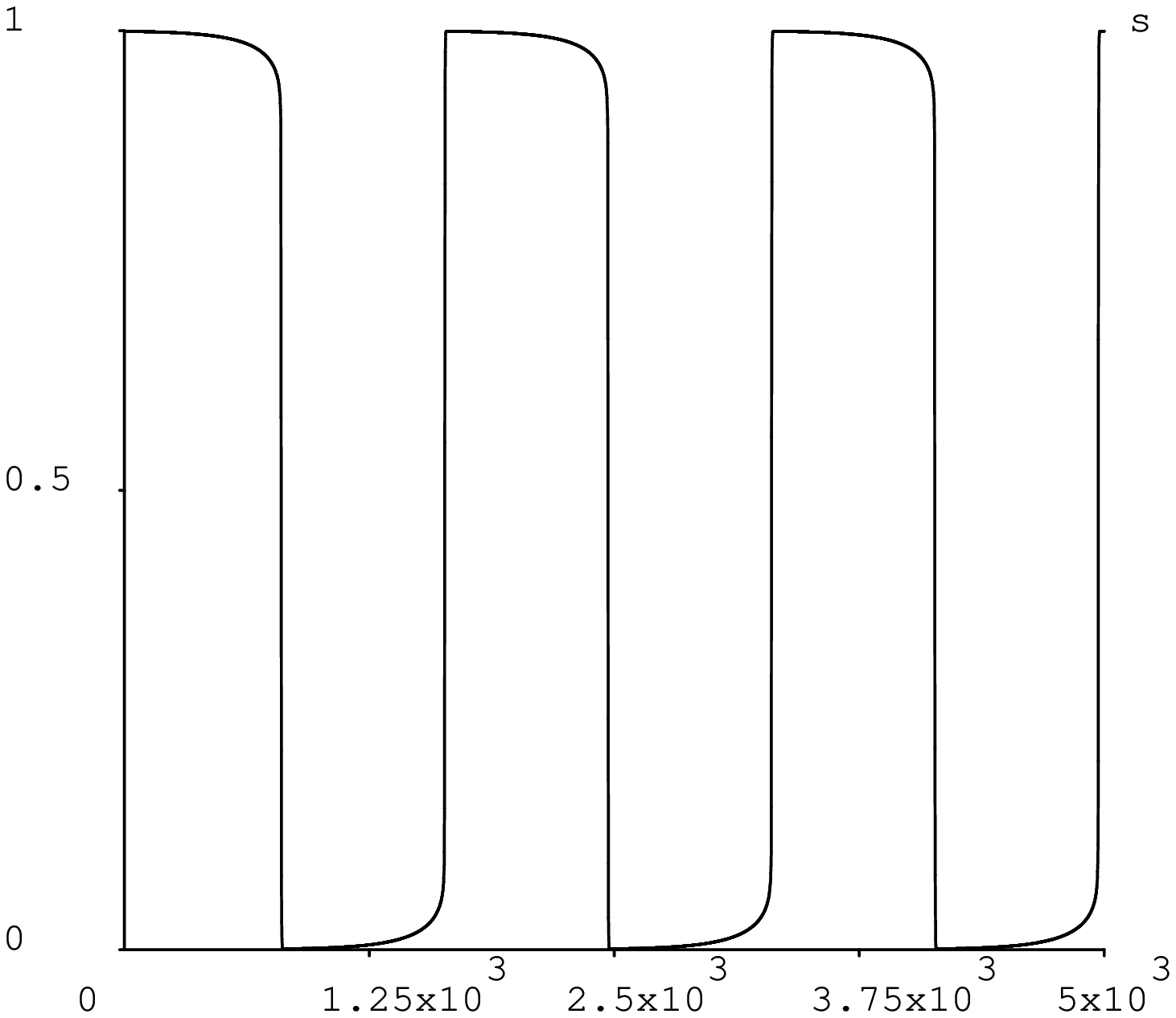,height=6cm} &  \psfig{file=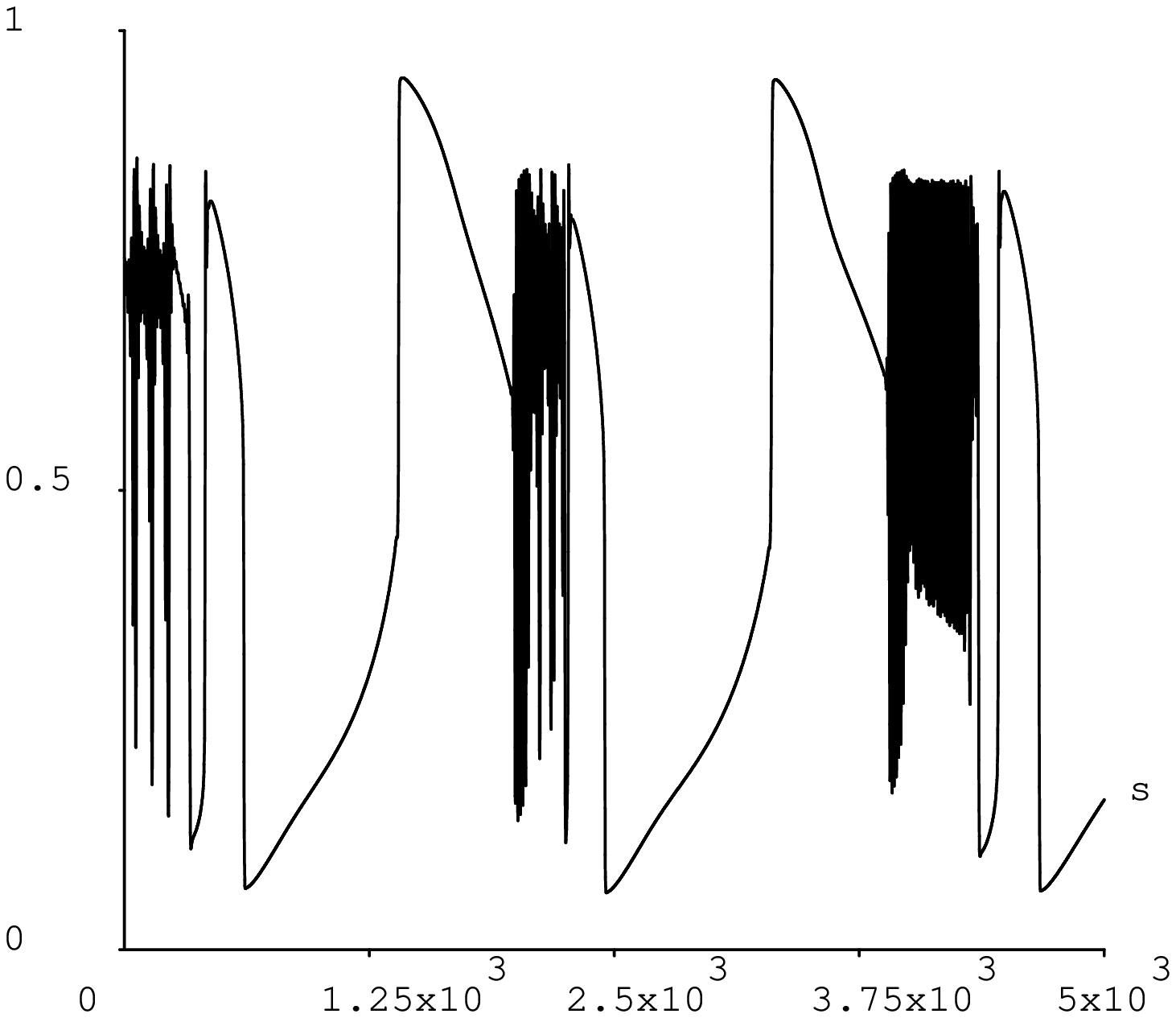,height=6cm} \\
    \psfig{file=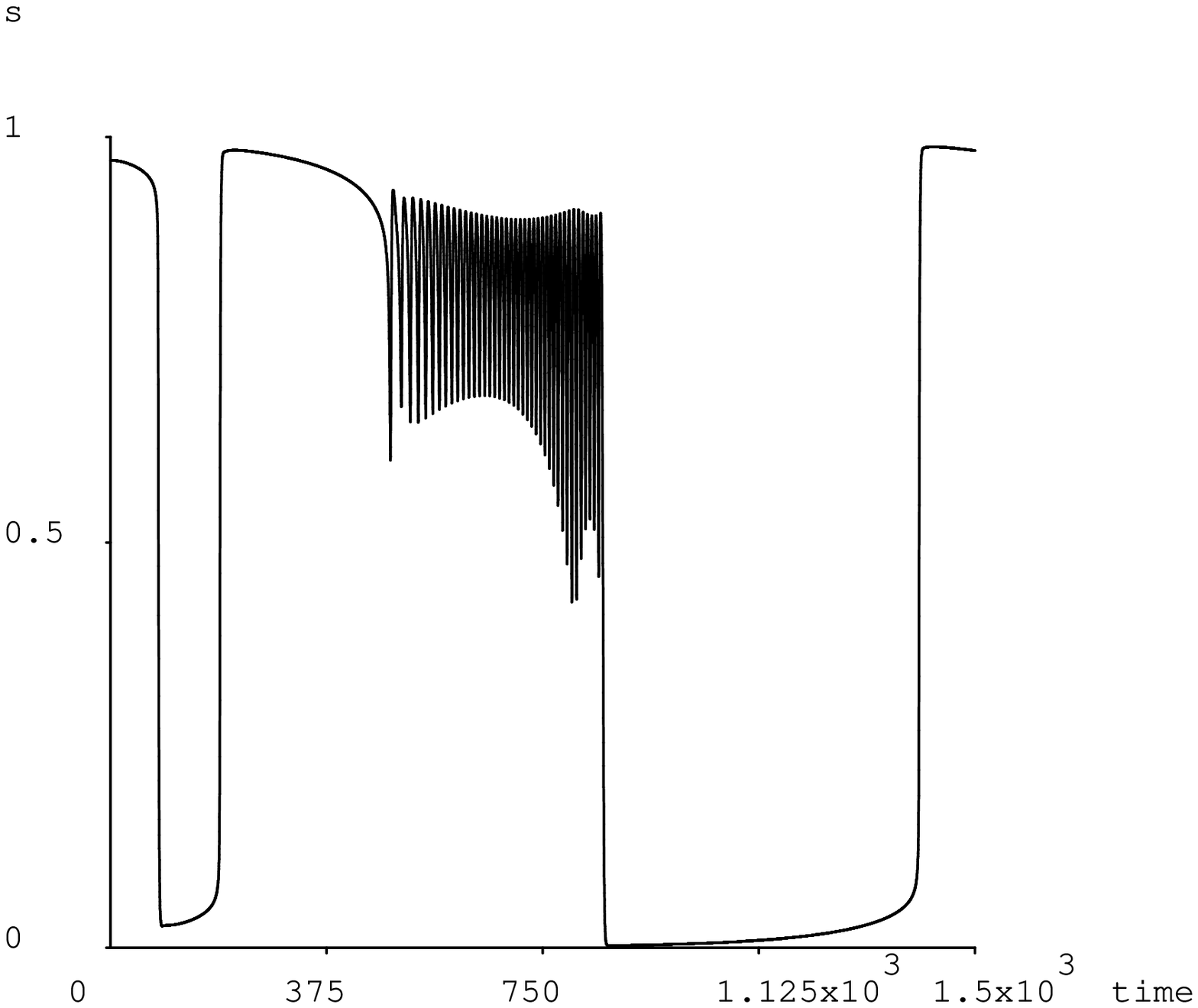,height=6cm} &  \psfig{file=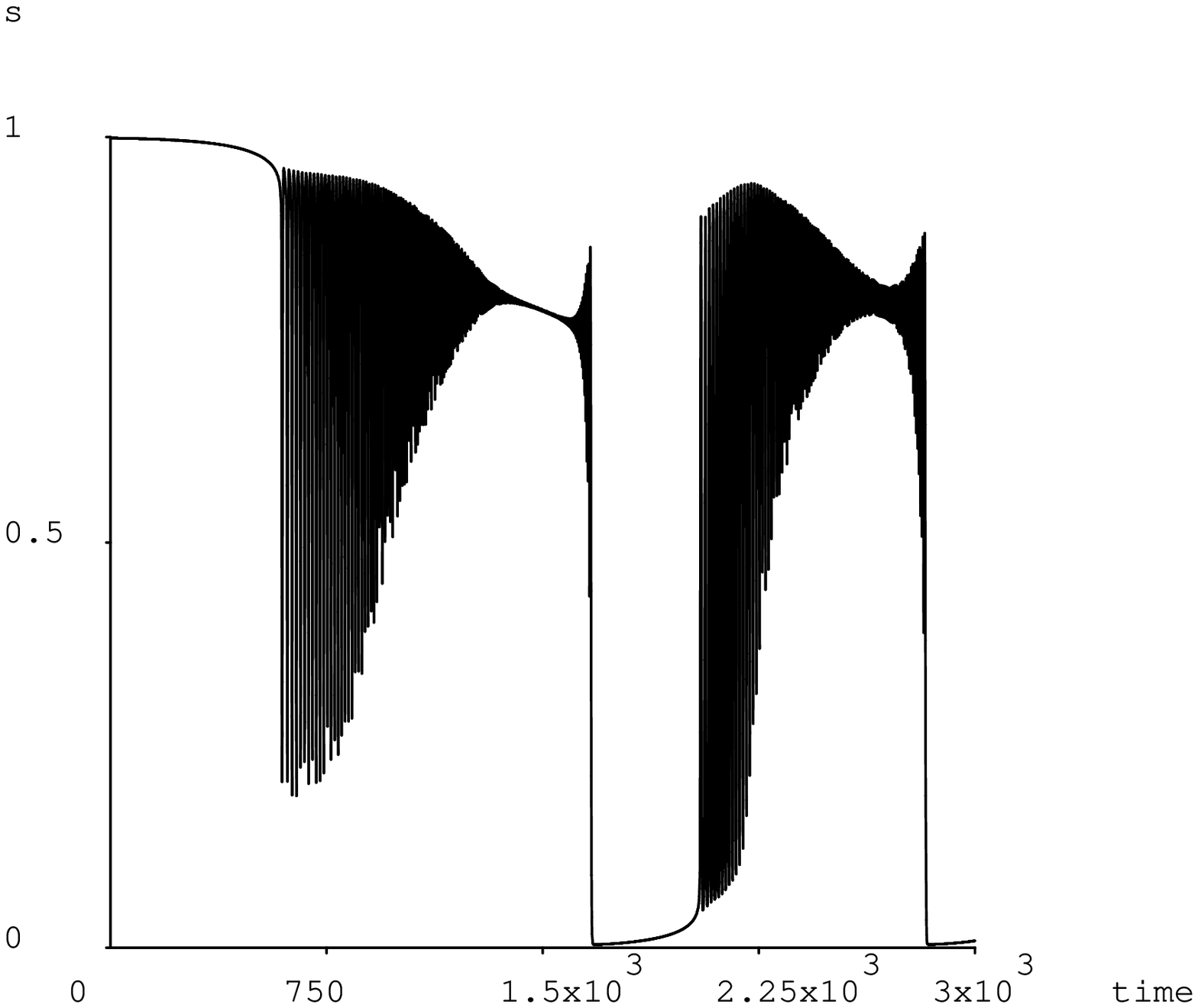,height=6cm} \\
    \psfig{file=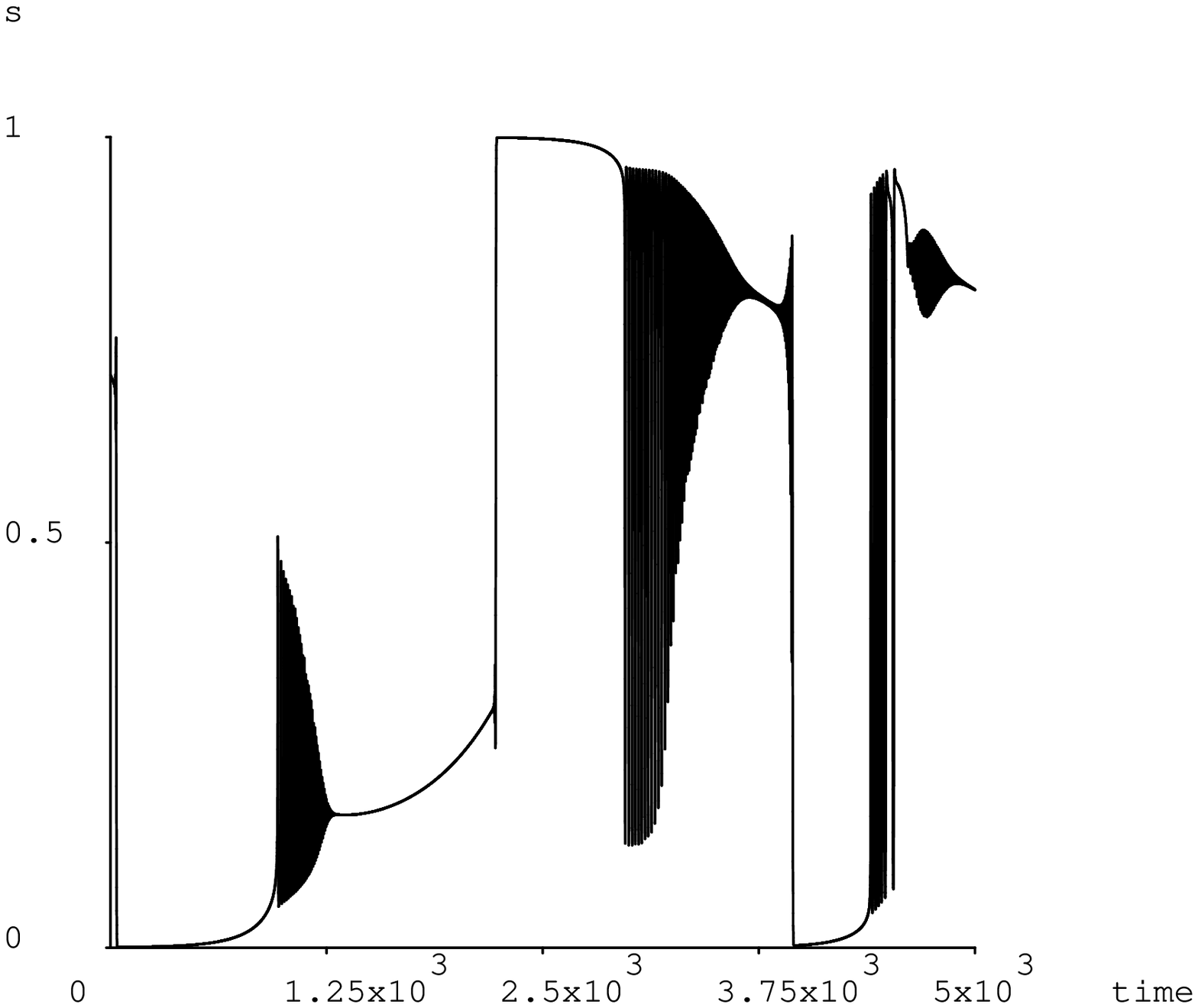,height=6cm} 
  \end{tabular}
\begin{quote}
Figure 8: Various behaviors of regulated full system
after it has reached critical surface (Figure 7).
Diagrams show $s(t)$ for three slightly different parameter settings (see text);
in all cases, the projection of the motion on the $(w\ee ,w\ie )$ plane
remains of small amplitude.
(a) Simple periodic attractor, point $G$ of Figure 7;
similar periodic attractors are reached for most parameter settings.
(b) Complex quasi-periodic attractor.
(c--e) Chaotic attractor;
for a given parameter setting,
three diagrams corresponding to different instants of time
and different lengths of time;
characteristic are the irregular transitions
between the high-activity, low-activity, and oscillatory phases.

\end{quote}

\end{document}